\pdfoutput=1

\PassOptionsToPackage{prologue,dvipsnames}{xcolor}

\documentclass[11pt]{article}

\usepackage[toc,page,header]{appendix}
\usepackage{minitoc}

\usepackage[final]{acl}

\usepackage{times}
\usepackage{latexsym}

\usepackage[T1]{fontenc}

\usepackage[utf8]{inputenc}

\usepackage{microtype}

\usepackage{inconsolata}

\usepackage{graphicx}

\usepackage{makecell}
\usepackage{multirow}
\usepackage{multicol}
\usepackage{caption}
\usepackage{subcaption}
\usepackage{wrapfig}
\usepackage{tablefootnote}
\usepackage{url}            
\usepackage{booktabs}       
\usepackage{amsfonts}       
\usepackage{nicefrac}       
\usepackage{microtype}      
\usepackage[export]{adjustbox}
\usepackage[many]{tcolorbox}
\renewcommand{\ttfamily}{\fontencoding{T1}\fontfamily{lmtt}\selectfont}

\usepackage{amsmath,bm}
\usepackage{amsfonts}
\usepackage{comment}
\usepackage{amssymb}
\usepackage{mathtools}
\usepackage{bbm}
\usepackage{enumitem}
\usepackage[dvipsnames]{xcolor}	
\usepackage{lipsum}  

\usepackage{pifont}
\newcommand{\cmark}{{\ding{51}}}%
\newcommand{\xmark}{{\ding{55}}}%
\definecolor{MycolorRed}{HTML}{c00000}
\definecolor{MycolorBlue}{HTML}{2e76b6}
\definecolor{MycolorGreen}{HTML}{71ad47}

\newcommand{\fig}{Fig.\,}

\makeatletter
\newcommand*{\da@rightarrow}{\mathchar"0\hexnumber@\symAMSa 4B }
\newcommand*{\da@leftarrow}{\mathchar"0\hexnumber@\symAMSa 4C }
\newcommand*{\xdashrightarrow}[2][]{%
  \mathrel{%
    \mathpalette{\da@xarrow{#1}{#2}{}\da@rightarrow{\,}{}}{}%
  }%
}
\newcommand{\xdashleftarrow}[2][]{%
  \mathrel{%
    \mathpalette{\da@xarrow{#1}{#2}\da@leftarrow{}{}{\,}}{}%
  }%
}
\newcommand*{\da@xarrow}[7]{%
  \sbox0{$\ifx#7\scriptstyle\scriptscriptstyle\else\scriptstyle\fi#5#1#6\m@th$}%
  \sbox2{$\ifx#7\scriptstyle\scriptscriptstyle\else\scriptstyle\fi#5#2#6\m@th$}%
  \sbox4{$#7\dabar@\m@th$}%
  \dimen@=\wd0 %
  \ifdim\wd2 >\dimen@
    \dimen@=\wd2 %
  \fi
  \count@=2 %
  \def\da@bars{\dabar@\dabar@}%
  \@whiledim\count@\wd4<\dimen@\do{%
    \advance\count@\@ne
    \expandafter\def\expandafter\da@bars\expandafter{%
      \da@bars
      \dabar@ 
    }%
  }%
  \mathrel{#3}%
  \mathrel{%
    \mathop{\da@bars}\limits
    \ifx\\#1\\%
    \else
      _{\copy0}%
    \fi
    \ifx\\#2\\%
    \else
      ^{\copy2}%
    \fi
  }%
  \mathrel{#4}%
}
\makeatother

\RequirePackage{xspace}
\makeatletter
\DeclareRobustCommand\onedot{\futurelet\@let@token\@onedot}
\def\@onedot{\ifx\@let@token.\else.\null\fi\xspace}

\def\eg{\emph{e.g}\onedot} 

\def\ie{\emph{i.e}\onedot}

\makeatother

\title{DocKD: Knowledge Distillation from LLMs for Open-World\\Document Understanding Models}

\author{
 \textbf{Sungnyun Kim\textsuperscript{1}\thanks{Equal contribution}\thanks{Work done at AWS AI Labs}\thanks{Corresponding author \href{mailto:ksn4397@kaist.ac.kr}{\texttt{ksn4397@kaist.ac.kr}}}},
 \textbf{Haofu Liao\textsuperscript{2}\footnotemark[1]},
 \textbf{Srikar Appalaraju\textsuperscript{2}},
 \textbf{Peng Tang\textsuperscript{2}},
 \textbf{Zhuowen Tu\textsuperscript{2}},
 \\
 \textbf{Ravi Kumar Satzoda\textsuperscript{2}},
 \textbf{R. Manmatha\textsuperscript{2}},
 \textbf{Vijay Mahadevan\textsuperscript{2}},
 \textbf{Stefano Soatto\textsuperscript{2}}
\\
\\
 \textsuperscript{1}KAIST AI\quad
 \textsuperscript{2}AWS AI Labs
}

\begin{document}
\maketitle

\doparttoc 
\faketableofcontents 

\begin{abstract}
    Visual document understanding (VDU) is a challenging task that involves understanding documents across various modalities (text and image) and layouts (forms, tables, etc.). This study aims to enhance generalizability of small VDU models by distilling knowledge from LLMs. We identify that directly prompting LLMs often fails to generate informative and useful data. In response, we present a new framework (called DocKD) that enriches the data generation process by integrating external document knowledge. Specifically, we provide an LLM with various document elements like key-value pairs, layouts, and descriptions, to elicit open-ended answers. Our experiments show that DocKD produces high-quality document annotations and surpasses the direct knowledge distillation approach that does not leverage external document knowledge. Moreover, student VDU models trained with solely DocKD-generated data is not only comparable to those trained with human-annotated data on in-domain tasks but also significantly excel them on out-of-domain tasks.
\end{abstract}
\section{Introduction}
\label{sec:intro}

Visual document understanding\,(VDU) requires extracting and analyzing both textual and non-textual information from a document. The textual information is usually obtained via optical character recognition\,(OCR), which only provides unstructured or na\"ively ordered text. The non-textual information is visually-rich, demanding a solution to directly process the document image. Earlier studies of VDU \cite{liu2007tableseer, hao2016table, soto2019visual} primarily focused on identifying certain parts of a document using heuristics or simple networks. Recent approaches \cite{huang2022layoutlmv3, tang2023udop} have shifted towards pretraining multi-modal document understanding models to address the model's comprehension of textual, visual, and layout features. However, the existing VDU methods are limited by training on a small-scale, curated document dataset, compromising the generalizability of VDU models to diverse documents. Thus, their performance heavily relies on the annotated training document set for downstream tasks.

\begin{figure}[!t]
    \centering
    \includegraphics[width=\linewidth]{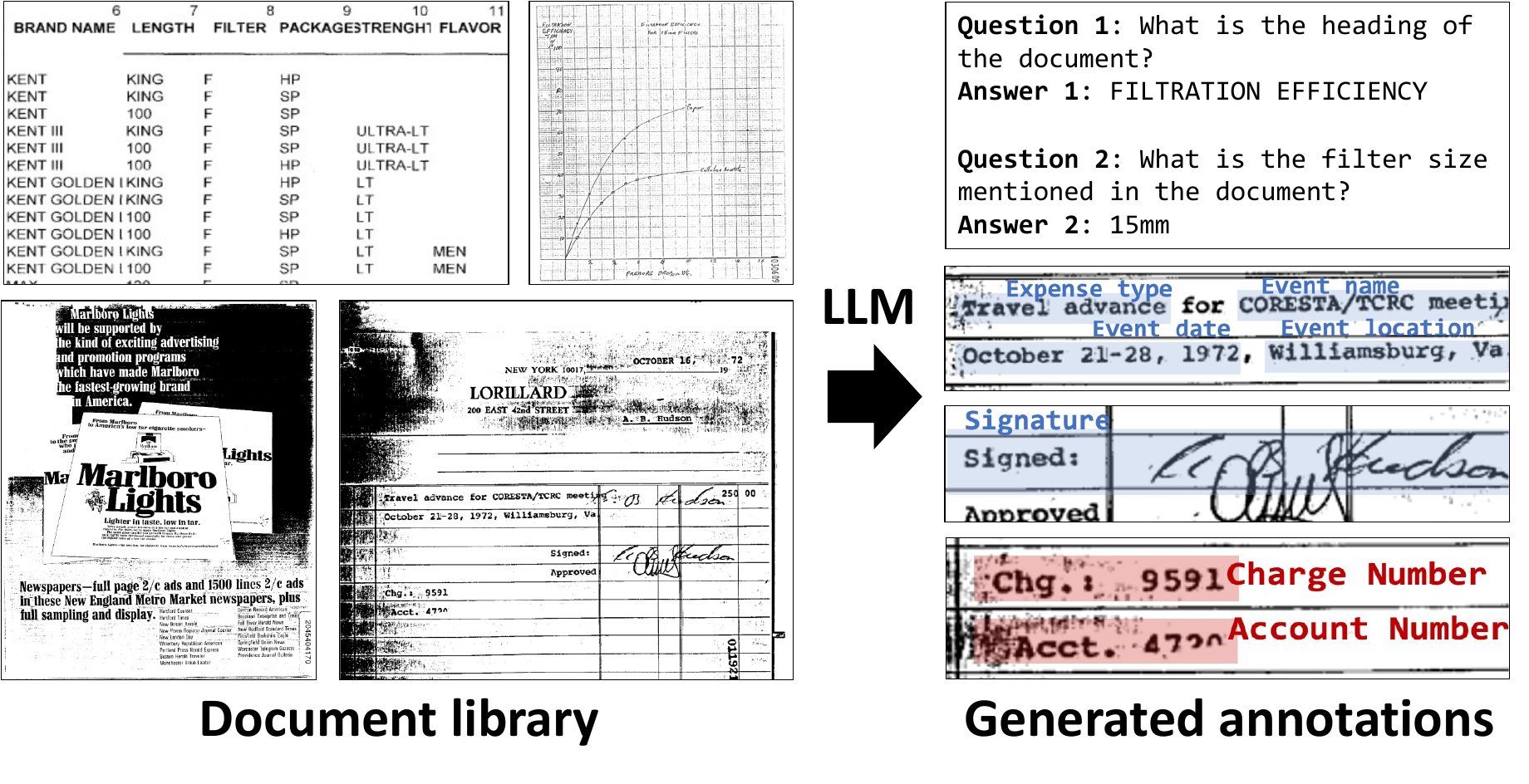}
    \vspace{-20pt}
    \caption{We leverage LLM to generate document annotations given the text extracted from a document image.}
    \label{fig:document_examples}
\end{figure}

In this study, we aim to improve the generalizability of VDU models by distilling knowledge from large language models (LLMs). In particular, we introduce an \textit{open-world document understanding} problem, where the model needs to address the downstream task with a broader scope of documents than covered by the available annotations.
LLMs, given instructions to elicit open-ended answers, can create rich and diverse annotations, as illustrated in Fig.\,\ref{fig:document_examples}.
For instance, we might instruct the LLM to ``\texttt{generate question-answer pairs from this document}'', along with document text extracted from OCR. However, this approach entails a critical challenge, since LLMs often struggle to comprehend unstructured OCR text \cite{wang2023latin}, leading to its generation of low-quality annotations. Moreover, there is a variety of non-textual information within documents which is not included in the LLM prompt.

To overcome these challenges, we present DocKD, \textit{a document knowledge distillation framework that leverages external document information to enhance LLM data generation}. In this framework, we extract various document elements (\emph{e.g.}, key-value pairs, layout, and descriptions) along with text and formulate a generation prompt for LLMs with this visual information. The LLM outputs then serve as annotations to train a small-scale VDU model.
While large multimodal models like GPT-4V \cite{2023GPT4VisionSC} are also recognized for their visual-language capabilities, they still lag behind state-of-the-art OCR systems \cite{fujitake2024dtrocr}, but LLMs that utilize well-structured OCR text excel in document processing and understanding. Thus, we employ LLMs aided with visual tools for data generation.

We demonstrate the efficacy of DocKD on three document understanding tasks: visual question answering, entity extraction, and classification. In each task, we introduce new tools for incorporating external document knowledge.
Our experiments reveal that DocKD allows student models to attain open document understanding abilities, generalizing to unseen documents, questions, entities, or categories. Our contributions are as follows:
\vspace{-5pt}
\begin{itemize}[label={$\circ$}, leftmargin=*]
\setlength\itemsep{0pt}
\item We introduce DocKD, a framework designed to facilitate VDU models for open-world document understanding. It boosts the generalizability of VDU models by leveraging LLMs and external document knowledge to generate training data. 

\item We demonstrate that DocKD surpasses direct knowledge distillation approach that relies solely on the LLM prompt tuning to generate data without document-specific knowledge.

\item In comparison to models trained with human-annotated data, student VDU models trained solely with DocKD-generated data achieve comparable performance on in-domain tasks and excel in addressing out-of-domain tasks. This showcases DocKD's potential to improve models for open-world documents understanding.

\end{itemize}
\section{Related Work}
\label{sec:related_work}

\paragraph{Document understanding models.}
Research in document intelligence\,\cite{liu2007tableseer, hao2016table, subramani2020survey, wang2022benchmark} has gained significant interest, developing machines to understand document contents and address associated tasks. Previous studies \cite{hong2020bros, wang2022lilt} have proposed document understanding models to improve the comprehension of multi-modality by integrating textual and layout information. 
These models later have evolved to incorporate visual information as well\,\cite{appalaraju2021docformer, gu2021unidoc, peng2022ernie}. These models are typically pretrained through self-supervised learning methods, such as word/line alignment\,\cite{appalaraju2023docformerv2, tang2023udop} or masked text/image modeling \cite{li2021selfdoc, huang2022layoutlmv3}. Subsequently, they undergo a fine-tuning phase for specific downstream tasks, which entails the manual annotation of documents. To facilitate the training of VDU models without the need for human labels, we propose knowledge distillation\,\cite{hinton2015distilling, gou2021knowledge} approach from LLMs.

\begin{figure*}[!t]
    \centering
    \includegraphics[width=\textwidth]{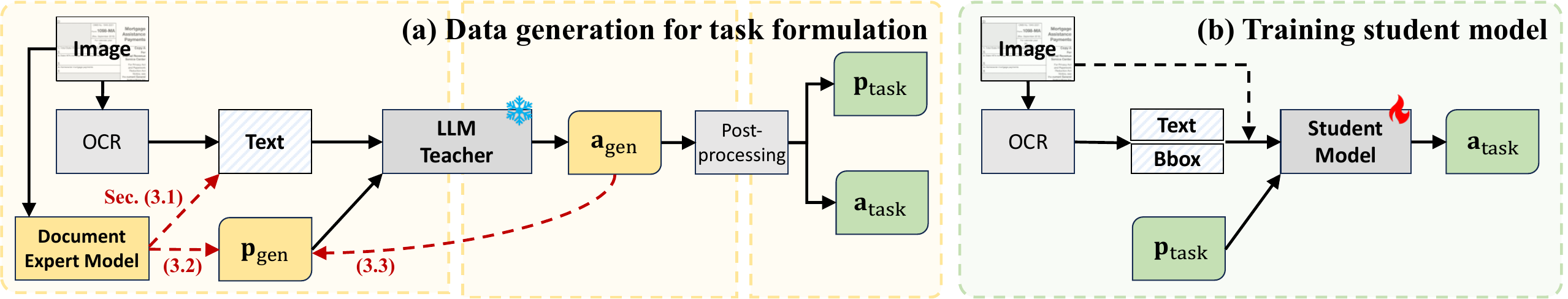}
    \caption{Overview of DocKD.
    \textbf{(a)} To prepare training data, we provide an LLM teacher with a generation prompt $\mathbf{p}_\text{gen}$ given the document text. LLM generates answers $\mathbf{a}_\text{gen}$ which are then converted into ($\mathbf{p}_\text{task}, \mathbf{a}_\text{task}$). We explore methods to inject external document knowledge (\textcolor{MycolorRed}{$\xdashrightarrow{}$}\!\!) into the document text or $\mathbf{p}_\text{gen}$ to obtain high-quality annotations.
    \textbf{(b)} We train a student VDU model using the generated task prompt and answer pairs ($\mathbf{p}_\text{task},\mathbf{a}_\text{task}$).
    }
    \label{fig:dockd}
\end{figure*}

\paragraph{Leveraging LLMs for data generation.}
Knowledge distillation\,(KD) from LLMs has been explored across various natural language processing tasks \cite{gu2023knowledge}. LLMs like GPT-3\,\cite{brown2020gpt3} are utilized for guided annotation of unlabeled data\,\cite{wang2021want, ding2022gpt, touvron2023llama2, vicuna2023} or for distilling reasoning capabilities \cite{magister2022teaching, hsieh2023distilling, zhu2023pad} which is then used to fine-tune smaller language models. Among these, targeted distillation\,\cite{jung2023impossible, zhou2023universalner} has demonstrated that identifying and amplifying the LLM's knowledge to a high-quality dataset enables student models to attain task-specific knowledge. It has the potential to make specialized language models that outperform in specific tasks, at the expense of generic performances\,\cite{fu2023specializing}.

In visual instruction tuning research\,\cite{li2023mimic, li2023blip, li2023m, liu2023llava, liu2023improved}, LLMs are employed to generate visual-language instruction-following data. For instance, LLaVA\,\cite{liu2023llava} is trained on the instruction-following dataset for conversation, description, and complex reasoning, created by prompting the LLM with bounding box coordinates of objects along with image captions. InstructBLIP\,\cite{dai2023instructblip} incorporates diverse tasks, such as image question generation and video question answering.
Closest to our work is the extension of visual instruction tuning to the domain of VDU, generating data with document-specific knowledge to fine-tune downstream models. \citet{wang2023layout} use layout-aware documents to answer given questions and fine-tune LLMs, and \citet{aubakirova2023patfig} generate captions for patent figures to fine-tune VLMs. The community has recently focused on directly improving the VDU performance of LLMs or LMMs by introducing new designs of encoding document images \cite{li2024enhancing, luo2024layoutllm, tanaka2024instructdoc, liu2024textmonkey}, which are closely related and complementary to our work that focuses on distilling knowledge from strong LLMs for VDU. 
Our work is the first to extract knowledge from LLMs for open document understanding tasks, exploring methods to inject visual document-specific knowledge into LLM and produce high-quality data for training VDU models.

\section{Document Knowledge Distillation}
\label{sec:method}

\paragraph{Problem formulation.}
Similar to prior work \cite{kim2022ocr, appalaraju2023docformerv2, tang2023udop}, we formulate document understanding problem under a sequence-to-sequence (seq2seq) generation framework. That is, we design a task-specific prompt $\mathbf{p}_\text{task}$ which asks a VDU model to solve the task and output an answer $\mathbf{a}_\text{task}$. DocKD involves an LLM teacher $f_\text{T}$ to generate these prompt and answer pairs.
Given an image of a document page, we apply a pre-built OCR engine to extract its words and word bounding boxes. For simplicity, we represent a document input as $\mathbf{d}$.

The overall pipeline of the DocKD approach is described in \fig\ref{fig:dockd}. In \fig\ref{fig:dockd}\,{(a)}, we first construct a generation prompt $\mathbf{p}_\text{gen}$ for the task. Then, given  $\mathbf{p}_\text{gen}$ and document text $\mathbf{d}_\text{text}$ as inputs, the LLM generates $\mathbf{a}_\text{gen}$, \emph{i.e.}, $f_\text{T}({\mathbf{d}_\text{text}}, \mathbf{p}_\text{gen}) \rightarrow \mathbf{a}_\text{gen}$. This can be readily parsed into $(\mathbf{p}_\text{task}, \mathbf{a}_\text{task})$ by post-processing.
Here, we can inject document-specific knowledge into the LLM inputs, so that it can better understand the document content and generate more accurate ($\mathbf{p}_\text{task}, \mathbf{a}_\text{task}$) pairs. In \fig\ref{fig:dockd}\,{(b)}, we train a student model $f_\text{S}$ to output an answer $\mathbf{a}_\text{task}$ given $\mathbf{d}$ and $\mathbf{p}_\text{task}$, \emph{i.e.}, $f_\text{S}(\mathbf{d}, \mathbf{p}_\text{task}) \rightarrow \mathbf{a}_\text{task}$.

We exemplify the application of our training pipeline on three document understanding tasks: visual question answering (VQA), entity extraction, and document classification.
To summarize each section, we leverage document knowledge by using the OCR linearization model to improve $\mathbf{d}_\text{text}$ (Sec.\,\ref{subsec:vqa}), using the key-value detection model to guide $\mathbf{p}_\text{gen}$ (Sec.\,\ref{subsec:entity_extraction}), and introducing the document description into $\mathbf{p}_\text{gen}$ for better class candidates (Sec.\,\ref{subsec:classification}).
Refer to Appx.\,\ref{appx:generative_prompts} for the full templates of $\mathbf{p}_\text{gen}$ in each task.

\subsection{Document VQA}
\label{subsec:vqa}

Document VQA \cite{borchmann2021due, mathew2021docvqa, mathew2022infographicvqa, van2023dude} is the task of answering questions about documents.
Given a document {$\mathbf{d}$} and a corresponding question-answer (QA) pair ($\textcolor{Green}{\mathbf{q}}, \textcolor{Orange}{\mathbf{a}}$), we design the task prompt as 
$\mathbf{p}_\text{task}=$\,``\texttt{Document:}\,{$\mathbf{d}_\text{text}$}. \texttt{Question:}\,$\textcolor{Green}{\mathbf{q}}$'', and $\mathbf{a}_\text{task}=$\,``\texttt{Answer:}\,$\textcolor{Orange}{\mathbf{a}}$''.  
To distill knowledge for a VDU model, we investigate a way to prompt LLMs to generate QA pairs from documents. 

\begin{figure}[!t]
    \centering
    \includegraphics[width=\linewidth]{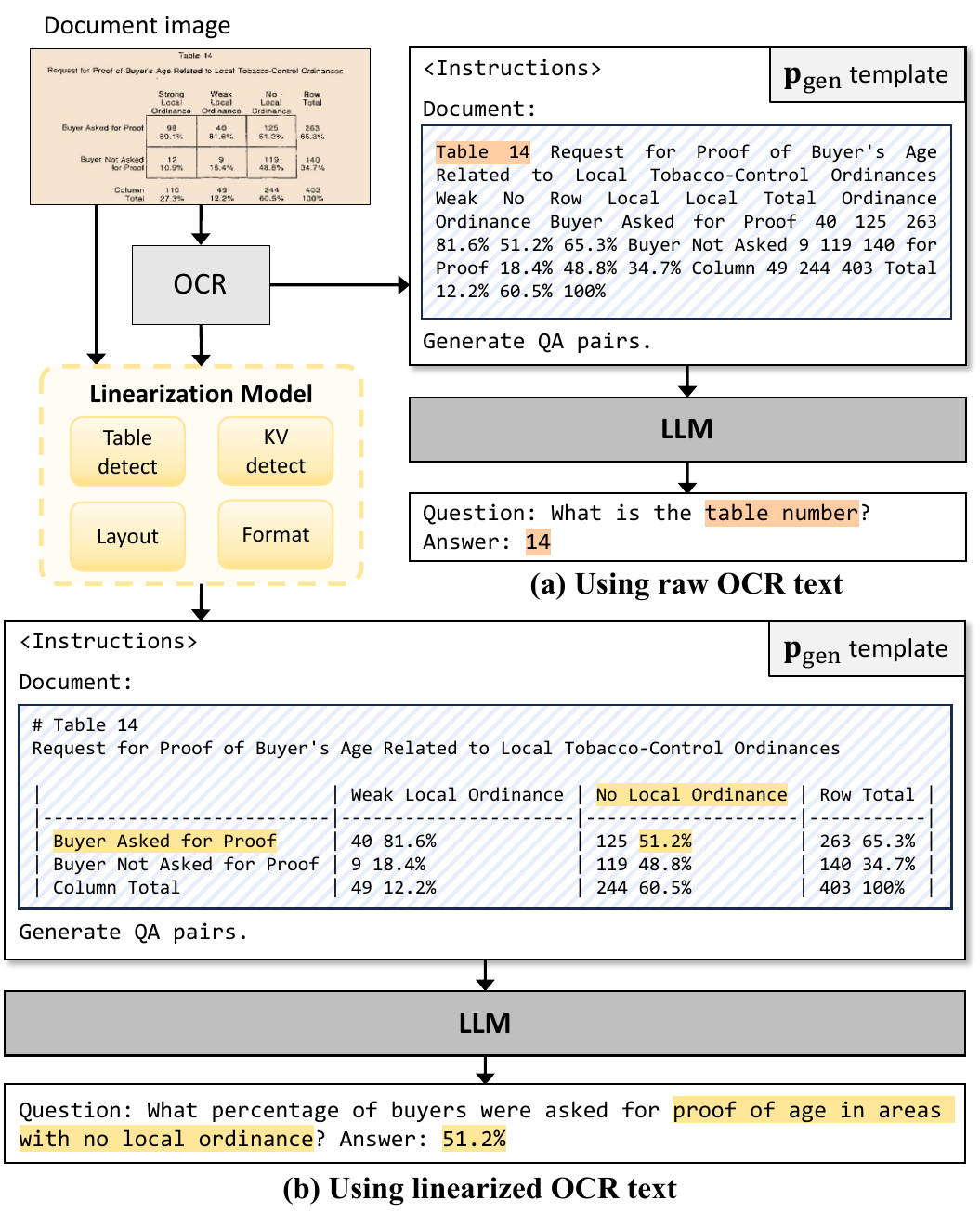}
    \vspace{-15pt}
    \caption{
    {(a)} When the input document text is in its raw OCR form, LLM produces simply extracted QA pairs.
    {(b)} When provided with linearized OCR text processed by a linearization model, LLM generates QA pairs that require visual layout knowledge to solve.}
    \label{fig:qa_generation}
\end{figure}

\paragraph{Designing QA generation task.}
Based on the OCR text as input context, we provide the LLM with a generation prompt $\mathbf{p}_\text{gen}$ to generate several QA pairs, as shown in \fig\ref{fig:qa_generation}\,{(a)}: 
$$f_\text{T}({\mathbf{d}_\text{text}}, \mathbf{p}_\text{gen}) \rightarrow \mathbf{a}_\text{gen} = \{(\textcolor{Green}{\mathbf{q}_1}, \textcolor{Orange}{\mathbf{a}_1}), (\textcolor{Green}{\mathbf{q}_2}, \textcolor{Orange}{\mathbf{a}_2}), \dots\}$$
We randomly select one question and its corresponding answer from $\mathbf{a}_\text{gen}$ and create ($\mathbf{p}_\text{task}, \mathbf{a}_\text{task}$) for training the student model.
We find that including an instruction into $\mathbf{p}_\text{gen}$ helps the teacher avoid creating low-quality QAs (\emph{e.g.}, duplicated questions or answers inconsistent with context) and enables us to control the generation output so that it can be easily parsed into ($\mathbf{p}_\text{task}, \mathbf{a}_\text{task}$).

We also note that $\mathbf{p}_\text{gen}$ instructs the LLM to output questions and answers \textit{together}, which we find facilitates the generation of accurate QA pairs.
Alternatively, we may ask the LLM to generate questions first and then answer them, which we observe that the generated questions are often difficult to answer, or the answers do not match the questions.

\paragraph{Introducing layout knowledge to OCR text.}
One limitation of the LLM's QA generation lies on its text-to-text framework, where it requires the text to be organized in a semantically meaningful order. However, OCR text is a simple sequence of words typically ordered by raster scanning, which ignores the important layout and structural information of document pages. Therefore, QAs generated from such text are usually less challenging and do not cover the spatial relationship between entities.

To ensure the LLM's awareness on the text layout, we replace the raw OCR text with spatially linearized OCR text, where we organize document text into a markdown style as displayed in \fig\ref{fig:qa_generation}\,{(b)}.
We use the linearization model inspired by \cite{peng2022ernie}, also extracting tables, key-value pairs, and layout information using Textract API\footnote{\url{https://aws.amazon.com/textract/}} which assists the conversion to markdown.
Interestingly, an LLM understands this markdown style; thus, the linearization model supplements document layout knowledge that is missing and helps the LLM to generate more diverse and higher-quality QAs. 
The student model trained with these QA pairs achieves notable VQA performances (Table\,\ref{tab:document_vqa}).
Refer to Appx.\,\ref{subsec:generated_qas} for the examples of generated QAs with raw or linearized OCR text.

\subsection{Entity Extraction}
\label{subsec:entity_extraction} 

\begin{figure*}[!t]
    \centering
    \includegraphics[width=\linewidth]{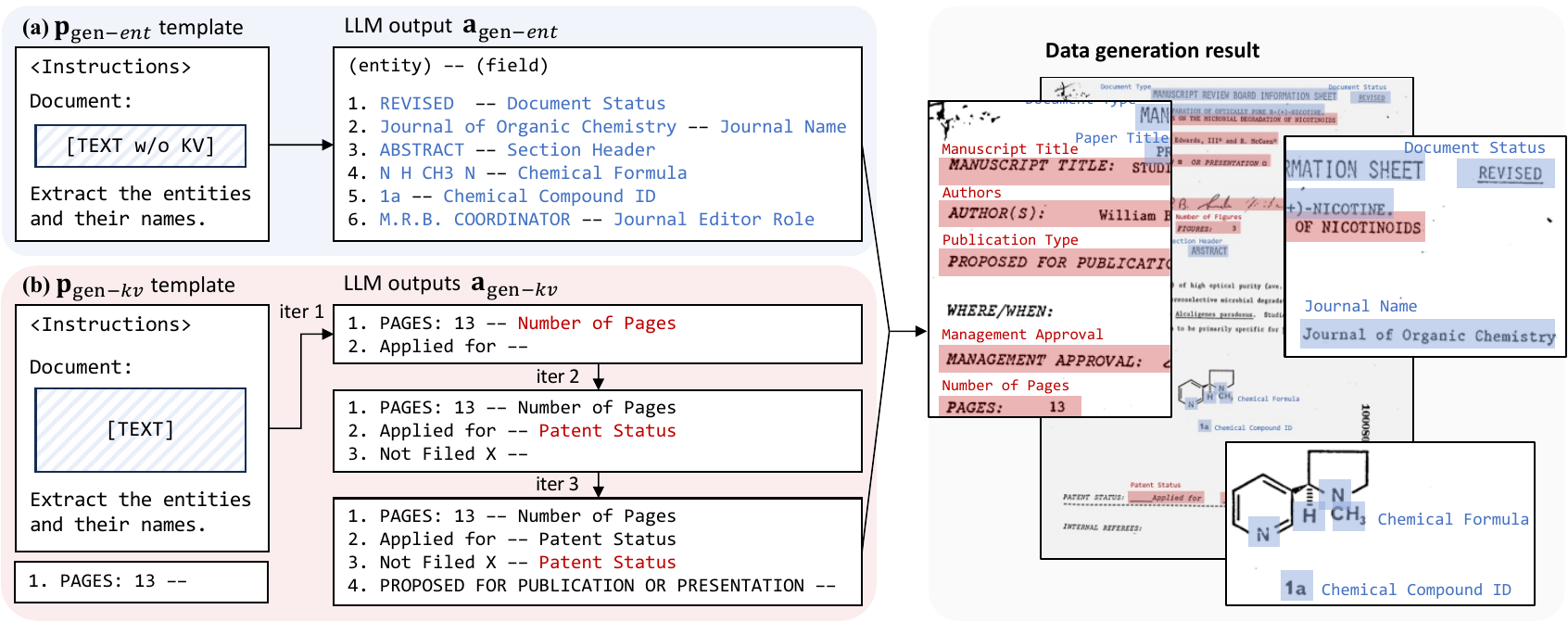}
    \caption{The templates on the left serve as input prompts to the LLM, for {(a)}\,generating non-KV entities and {(b)}\,naming KV entities, respectively. For {(b)}, in the iteration $n$, the $n$-th KV entity is provided as input as well as the output from the previous iteration. On the right, we show the result of generated entities and field names, with blue boxes representing non-KV entities and red boxes representing KV entities.}
    \label{fig:entity_extraction}
\end{figure*}

Entity extraction aims to identify entities in the document that matches a given field name. 
Similar to the VQA task, we convert this task into a seq2seq form. 
For each field name \textcolor{Green}{$\mathbf{f}$} and the corresponding entity \textcolor{Orange}{$\mathbf{e}$}, $\mathbf{p}_\text{task}=$\,``\texttt{Document:} {$\mathbf{d}_\text{text}$}. \texttt{Question: what are entities of $<\!\textcolor{Green}{\mathbf{f}}\!>$?}'' and $\mathbf{a}_\text{task}=$\,``\texttt{Answer:} \textcolor{Orange}{$\mathbf{e}$}''.

The challenge of this task lies in that we do not know which field will be queried for a new document. Thus, we should generate as many diverse fields as possible for different kinds of entities, and train the entity extraction model to link those fields to the entities. Indeed, LLMs are known to be proficient at the entity recognition task \cite{li2019unified, wang2023gpt-ner} and can even identify their names \cite{zhou2023universalner}.

\vspace*{-3pt}
\paragraph{Designing entity generation task.} To generate data for entity extraction, we prompt LLMs to exhaustively extract any entities present in a document. We design an entity extraction prompt $\mathbf{p}_{\text{gen-}\textit{ent}}$ and send it together with the document text $\mathbf{d}_\text{text}$ as the inputs to an LLM, which then outputs a list of entities along with their field names:
$$f_\text{T}(\mathbf{d}_\text{text}, \mathbf{p}_{\text{gen-}\textit{ent}}) \!\rightarrow\! \mathbf{a}_{\text{gen-}\textit{ent}} \!=\! \{\!(\textcolor{Green}{\mathbf{f}_1}, \textcolor{Orange}{\mathbf{e}_1}), (\textcolor{Green}{\mathbf{f}_2}, \textcolor{Orange}{\mathbf{e}_2}),\!... \}$$
where $\mathbf{f}_i$ is a generated field name for the $i$-th entity $\mathbf{e}_i$. We find that LLMs are able to capture a group of words into a single entity and generate a field based on the context, as observed in \fig\ref{fig:entity_extraction}\,{(a)}.

\vspace*{-3pt}
\paragraph{Introducing KV entity knowledge to $\mathbf{p}_{\text{gen}}$.}
Although LLMs can identify entities from documents to a certain extent, we notice that they are unable to sufficiently enumerate the entities. They tend to list mostly the major ones, especially when there are many potential entities in the document, and fail to identify diverse types. To help LLMs to enumerate them, we propose to leverage a document expert model that extracts key-value (KV) pairs from documents. KV pairs are frequently found in documents, \emph{e.g.}, the entity ``\texttt{Name:\,XYZ}'' is composed of a key ``\texttt{Name:}'' and a value ``\texttt{XYZ}''.

We detect all KV pairs using an external KV detection model, and send the detected KV pairs to LLMs to obtain their field names. Because there exist multiple KV pairs, we iteratively present each KV entity line by line to the LLM, with the previous line's output appended (refer to \fig\ref{fig:entity_extraction}\,{(b)}):
$$f_\text{T}(\mathbf{d}_\text{text}, \mathbf{p}_{\text{gen-}\textit{kv}}, (\mathbf{f}_i, \mathbf{e}_i)_{1:n}, \textcolor{Orange}{\mathbf{e}_{n+1}}) \!\rightarrow\! \mathbf{a}_{\text{gen-}\textit{kv}} \!= \textcolor{Green}{\mathbf{f}_{n+1}}$$
where $\mathbf{f}_{n+1}$ is a field name for the KV entity $\mathbf{e}_{n+1}$, as result of the $(n+1)$-th generation.
This way, we make the LLM focus on the field generation only for the current KV entity. In addition, it has access to previous generated outputs, so if there are similar entities given, it can assign the same field.

Note that we do not eliminate the entity generation process by $\mathbf{p}_{\text{gen-}\textit{ent}}$. Not all entities are detected by the KV detection model, so it is still required to extract non-KV entities. Hence, when generating non-KV entities, we provide the OCR text in which all KV entities are removed.

\subsection{Document Classification}
\label{subsec:classification}

We formulate a classification task within a seq2seq framework so that a VDU model can generalize to any novel classes.
Specifically, we design the input prompt as
$\mathbf{p}_\text{task}=$ ``\texttt{Document:}\,{$\mathbf{d}_\text{text}$}. \texttt{Question:} \texttt{what is the class of this document? choose from the following:\,\{\textcolor{Green}{\textsl{candidate list}}\}}'', and correspondingly, $\mathbf{a}_\text{task}=$ ``\texttt{Answer:} \texttt{\textcolor{Orange}{\textsl{class label}}}''. The candidate list contains document class labels, including the answer class. We collect the LLM-generated labels to fill out the prompt without human annotations.

\vspace*{-3pt}
\paragraph{Designing document class generation task.}
We generate candidates of class labels that can further be used to formulate a downstream classification task.
For this, we need two types of generation prompts.
$\mathbf{p}_\text{gen-\textit{pos}}$ is used to generate candidates of a given document's type, and we call this output list \textit{positive labels} that may be used as an answer.
In order to build a classification task, we not only need the document types that match the given document but also the candidate types that do not match the document.
LLM is instructed with $\mathbf{p}_\text{gen-\textit{neg}}$ to suggest these types, which we call \textit{negative labels}.

\vspace*{-3pt}
\paragraph{Introducing knowledge from $\mathbf{a}_{\text{gen}}$ to $\mathbf{p}_{\text{gen}}$.}
We notice that when an LLM is directly prompted to predict document classes, it frequently generates class labels that are overly general, resulting in low diversity. To address this, we incorporate document descriptions to $\mathbf{p}_{\text{gen}}$ which we find can facilitate LLMs to better summarize a document and generate more diverse class labels.

LLM is instructed with $\mathbf{p}_\text{gen-\textit{desc}}\!=$\,``\texttt{Describe this document in one sentence}''. The output document description $\mathbf{a}_\text{gen-\textit{desc}}$ is then appended to the generation prompt for positive labels.
This strategy makes the positive labels more diverse and detailed, \emph{e.g.}, \texttt{\textsl{letter}}\,$\rightarrow$\,\texttt{\textsl{consumer letter}}.
Subsequently, we also use the output positives in the negatives generation prompt, in order to avoid generating labels that are similar to the positives. We summarize the generation steps as follows:
\begin{align*}
    &\text{(1) description:}~f_\text{T}(\mathbf{d}_\text{text}, \mathbf{p}_\text{gen-\textit{desc}}) \!\rightarrow\! \mathbf{a}_\text{gen-\textit{desc}}, \\
    &\text{(2) positives:}~f_\text{T}(\mathbf{d}_\text{text}, \mathbf{p}_\text{gen-\textit{pos}}, \mathbf{a}_\text{gen-\textit{desc}}) \!\rightarrow\! \mathbf{a}_\text{gen-\textit{pos}}, \\
    &\text{(3) negatives:}~f_\text{T}(\mathbf{d}_\text{text}, \mathbf{p}_\text{gen-\textit{neg}}, \mathbf{a}_\text{gen-\textit{pos}}) \!\rightarrow\! \mathbf{a}_\text{gen-\textit{neg}}.
\end{align*}
While this approach does not directly leverage visual information, it adopts a similar strategy to the chain-of-thought reasoning\,\cite{wei2022chain, hsieh2023distilling} that encourages better outputs by prompting the instruction steps to LLMs.

\paragraph{Candidate list formulation.}
We select one positive label the list $\mathbf{a}_\text{gen-\textit{pos}}$, as an answer. For other non-answer candidates, we randomly sample a few from $\mathbf{a}_\text{gen-\textit{neg}}$. We train the model to choose one among the \texttt{\{\textsl{positive}\,+\,\textsl{negatives}\}} list.
In addition, the generated description $\mathbf{a}_\text{gen-\textit{desc}}$ is appended to each positive label to give a hint about the class.
We also gather all unique negative classes and use the LLM to produce descriptions for these types, which are also appended to the labels. Refer to Appx.\ref{appx:generative_prompts_for_cls} for the prompt we used based on this.

\section{Experiments and Results}
\label{sec:experiments}

\subsection{Implementation Details}

\begin{table*}[!t]
    \centering
    \footnotesize
    \addtolength{\tabcolsep}{2pt}
    \resizebox{.9\linewidth}{!}{
    \begin{tabular}{l|l|cc|cc|cc}
    \multicolumn{2}{c}{} & \multicolumn{2}{c}{(a) VQA} & \multicolumn{2}{c}{(b) Entity extraction} & \multicolumn{2}{c}{(c) Classification} \\[2pt]
    model & size & \!\!val ANLS\!\! & val EM & test F1 & \!\!test ANLS\!\! & \!\!test mAcc\!\! & \!\!test mAcc$^\star$\!\! \\
    \Xhline{2\arrayrulewidth}
    \multicolumn{5}{l}{\textcolor{gray}{\textit{LLM zero-shot prediction}}} \\
    \hline
    Flan-T5$_{\text{large}}$ \cite{chung2022flan} & 750M & 59.6 & 48.8 & 0.90 & 2.57 & 46.7 & 54.0 \\
    Flan-T5$_{\text{XXL}}$ \cite{chung2022flan} & 11B & {70.4} & {60.0} & 21.2 & 24.1 & 52.0 & 58.1 \\
    LLaVA-1.5~\cite{liu2023improved} & 13B & 49.0 & 37.3 & 9.12 & 5.20 & 36.1 & 43.3 \\
    Vicuna-1.3 \cite{vicuna2023} & 33B & 62.4 & 51.9 & 24.3 & 27.6 & 48.4 & 57.7 \\
    Falcon \cite{almazrouei@falcon40b} & 40B & 72.4 & 62.7 & {48.5} & {38.7} & 37.9 & 43.3 \\
    \Xhline{2\arrayrulewidth}
    \multicolumn{5}{l}{\textcolor{gray}{\textit{VDU models trained with \textcolor{Tan}{\textbf{only generated}} data}}} \\
    \hline
    Flan-T5$_{\text{large}}$ + KD & 750M & 70.4 & 59.4 & 24.4 & 56.3 & 52.3 & 59.8 \\
    Flan-T5$_{\text{large}}$ + \textbf{DocKD} & 750M & 72.9 & 62.7 & 55.9 & 66.1 & 57.0 & 71.7 \\ 
    DocFormerv2$_{\text{large}}$ + KD & 750M & 76.9 & 67.4 & 30.2 & 51.8 & 58.6 & 69.0 \\
    DocFormerv2$_{\text{large}}$ + \textbf{DocKD} & 750M & {\textbf{81.0}} & {\textbf{71.9}} & \textbf{61.5} & \textbf{68.7} & \textbf{62.4} & \textbf{73.9} \\
    \end{tabular}
    }
    \vspace*{-5pt}
    \caption{
    Document understanding results for LLMs and student VDU models. Note that none of these models were trained with human-labeled annotations.
    {(a)} DocVQA validation performance. 
    KD baseline uses raw OCR text for the QA generation, while DocKD uses linearized OCR text.
    {(b)} Entity extraction performance on CORD\,(F1) and DeepForm\,(ANLS).
    KD baseline generates entities without KV detection.
    {(c)} RVL-CDIP test accuracy. For DocKD, both class labels and descriptions are generated. mAcc$^\star$ measures the mean accuracy excluding four ambiguous categories: memo, filefolder, handwritten, and presentation.
    }
    \label{tab:document_vqa}
\end{table*}

\paragraph{Models.}
We compare the DocKD performance with the plain KD approach, na\"ively using $\mathbf{d}_\text{text}$ and $\mathbf{p}_\text{gen}$ without external document knowledge, as a prompt engineering baseline.
{By default, we use Claude-2 \footnote{\url{https://www.anthropic.com/index/claude-2}} as a teacher LLM and DocFormerv2$_\text{large}$ \cite{appalaraju2023docformerv2} as a student VDU model, while partially using DocFormerv2$_\text{base}$ to facilitate more efficient analysis.}
The training procedure of DocFormerv2\,(DFv2) closely follows that of the original paper, where it jointly encodes document image, OCR text, and bounding boxes. The provided query ($\mathbf{p}_\text{task}$) is appended to the text ($\mathbf{d}_\text{text}$), and the decoder outputs the target answer ($\mathbf{a}_\text{task}$).

For comparison, we also employ Flan-T5$_\text{large}$ \cite{chung2022flan} as a student language-only model, since the DFv2 structure is based on T5 \cite{raffel2020t5}.
To provide a base comparison for each task, we additionally present the zero-shot performance of instruction-tuned LLMs \cite{chung2022flan, almazrouei@falcon40b, vicuna2023} and a vision-language multi-modal foundation model \cite{liu2023improved}.

\paragraph{Datasets.}
For the LLM's data generation, we use a randomly sampled subset of Industry Document Library\,(IDL, \citet{idl}) as unannotated document images.
To accurately evaluate the open-world capabilities, we have removed all IDL documents that overlap with any of our downstream task datasets and excluded them from the data generation phase.
For the evaluation datasets and metrics, we use DocVQA \cite{mathew2021docvqa} validation set in the document VQA task, measured by ANLS (average normalized Levenshtein similarity) \cite{biten2019anls} and EM (exact match). In the entity extraction, we use two datasets, CORD \cite{park2019cord} and DeepForm \cite{borchmann2021due}, evaluated by entity-level F1 score and ANLS, respectively. In the classification task, we use RVL-CDIP \cite{harley2015rvlcdip} test set, evaluated by the mean accuracy over 16 document categories. Refer to Appx.\,\ref{appx:dataset_specifications} for more details on each dataset.

\subsection{Evaluation on Open-World Document Understanding Tasks}
\label{subsec:evaluation}

\paragraph{Document VQA.}
Claude-2 generates QAs from randomly sampled 100K IDL documents. We prompt Claude-2 to generate three QA pairs per document sample, and the trained student model is evaluated on DocVQA \cite{mathew2021docvqa}.
Table\,\ref{tab:document_vqa}\,{(a)} summarizes the DocVQA performances of the distilled students as well as the LLMs, where none of these models have been trained on human annotations for the document VQA task.
We confirm that knowledge-distilled student models can effectively answer document questions, being comparable with much larger-size language models.

Compared to the plain KD with raw OCR text, DocKD significantly enhances the performance up to 81.0\% ANLS. This result is comparable to using human-labeled annotations (refer to Sec.\,\ref{subsec:analysis}), which implies the high quality of generated data.
Furthermore, the performance gain is greater with DFv2\,(vision\,+\,language) than Flan-T5\,(language), which shows that the linearization model supplements informative visual knowledge.

\paragraph{Entity extraction.}
For generating the entities with KV detection, we need documents with rich key and value information. Such documents are frequently found from forms or invoices. Thus, instead of using IDL, we use the invoices subset of RVL-CDIP \cite{harley2015rvlcdip} for entity generation, sampling 5K documents. Table\,\ref{tab:document_vqa}\,{(b)} demonstrates that if the data generation does not involve the KV detection model but only exploits the entity generation prompt $\mathbf{p}_{\text{gen-}\textit{ent}}$, the LLM produces low-quality entities and field names, leading to the subpar performance of the student models.

\vspace*{-3pt}
\paragraph{Document classification.}
We sample 50K documents from IDL to generate class labels. For each document sample, Claude-2 generates one-sentence description, three positive labels, and ten negative labels. Table\,\ref{tab:document_vqa}\,(c) shows that our distillation framework enables the student model to classify novel documents, removing the need to pre-define categories or collect annotated documents to train a classification model. In addition, we find that DocKD's description generation induces more knowledge on documents compared to the plain KD, improving the accuracy by large margin: 58.6\%\,$\rightarrow$\,62.4\% mAcc.

\begin{figure}[!t]
     \centering
     \small
     \includegraphics[width=.88\linewidth]{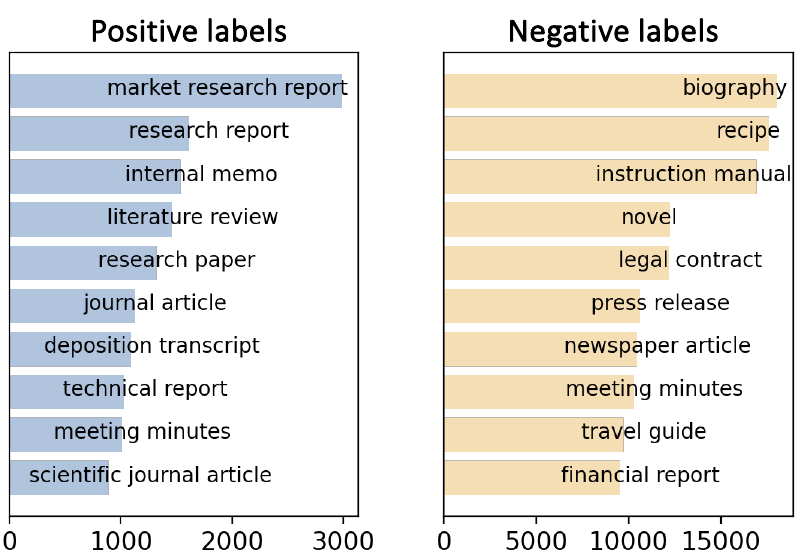}
     \vspace*{-7pt}
     \caption{Top-10 frequently generated document class labels from IDL \cite{idl}.}
     \label{fig:label_stats}
\end{figure}

\fig\ref{fig:label_stats} shows the spectrum of generated class labels from the IDL documents. After filtering out invalid labels (\emph{e.g.}, too long or outliers), it amounts to 49.9K unique positive labels and 10.5K unique negative labels. Before introducing the description generation, we had 17.2K unique positives, implying that the provision of description contributes to increasing the label diversity.

\begin{table}[!t]
    \centering
    \small
    \addtolength{\tabcolsep}{-2pt}
    \resizebox{\linewidth}{!}{
    \begin{tabular}{l|l|c|c|c|c}
    \multicolumn{2}{c|}{\,} & \!DocVQA\! & CORD & \!DeepForm\! & \!RVL-CDIP\! \\
    teacher & student & \scriptsize val ANLS & \scriptsize test F1 & \scriptsize test ANLS & \scriptsize test mAcc \\
    \Xhline{2\arrayrulewidth}
    Falcon-40B & DFv2$_\text{base}$ & 68.6 & 55.1 & 48.5 & 54.7 \\
    Falcon-180B & DFv2$_\text{base}$ & 71.3 & 59.8 & 62.0 & 53.8 \\
    Claude-2 & DFv2$_\text{base}$ & \textbf{77.2} & \textbf{60.2} & \textbf{64.2} & \textbf{61.9} \\
    \hline
    Falcon-40B & DFv2$_\text{large}$ & 74.9 & 59.8 & 61.2 & 55.6 \\
    Falcon-180B & DFv2$_\text{large}$ & 76.8 & \textbf{66.6} & 64.5 & 58.5 \\
    \textbf{Claude-2} & \textbf{DFv2$_\text{large}$} & \textbf{81.0} & {61.5} & \textbf{68.7} & \textbf{62.4} \\
    \end{tabular}
    }
    \vspace*{-5pt}
    \caption{We compare the Claude-2 teacher with Falcon-40B and Falcon-180B teacher models, and the DFv2$_\text{large}$ (750M) and DFv2$_\text{base}$ (232M) student models.}
    \label{tab:small_teacher_student}
\end{table}

\paragraph{Smaller teacher and student models.} 
Table\,\ref{tab:small_teacher_student} presents the result with a smaller teacher, Falcon-40B \cite{almazrouei@falcon40b}, and a smaller student, DFv2$_\text{base}$.
We find that smaller teacher and student models can degrade the data generation quality and task performances.
In contrast, larger and stronger teacher models like Claude-2 or Falcon-180B \cite{falcon180b} can generate better data, leading to the highest task performances. For instance, Claude-2 better understands the linearized OCR text than Falcon-40B does, so it generates diverse and accurate QAs from the layout-aware text. Refer to Appx.\,\ref{appx:generated_annotations} for comparisons between different teacher models.

\begin{figure}[!t]
    \centering
    \includegraphics[width=\linewidth]{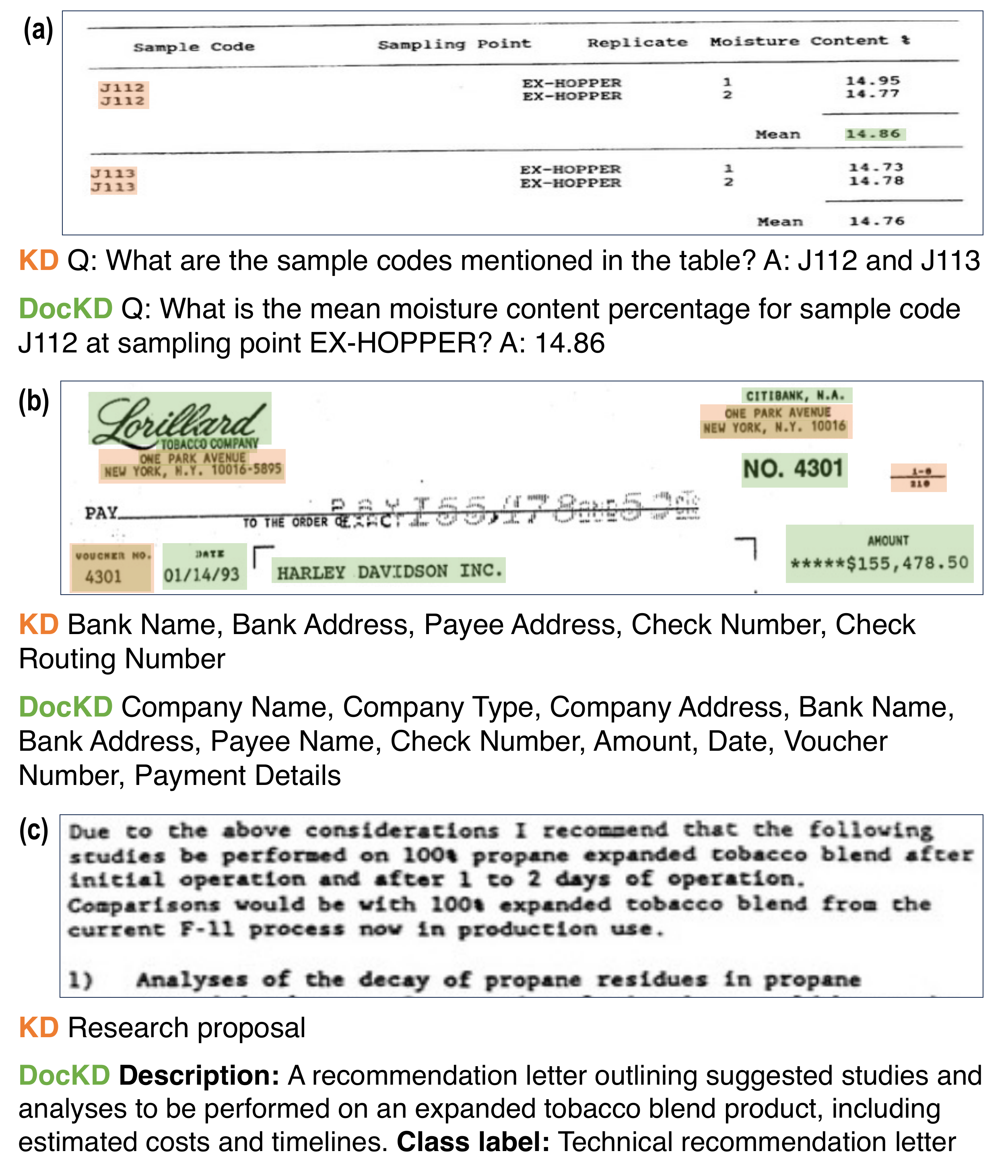}
    \vspace{-20pt}
    \caption{Comparison between data generated by KD and DocKD: (a)\,document VQA, (b)\, entity extraction, and (c)\,document classification.}
    \label{fig:annotation_examples}
\end{figure}

\paragraph{Visualization and statistics of generated data.}
Fig.\,\ref{fig:annotation_examples} visualizes some sample data generated by using KD and DocKD respectively. For document VQA, DocKD generates more challenging QA pairs that requires understanding the structure of the table. In Fig.\,\ref{fig:annotation_examples} (a), the question generated by DocKD requires understanding the relationship between ``mean'', ``moisture content \%'', ``sample code'' and ``sample point''. For entity extraction, we show a common example in Fig.\,\ref{fig:annotation_examples} (b) where we list the entity names extracted by KD and DocKD. We see that DocKD is able to capture significantly more entities than KD. For document classification, we note that DocKD generates a document description which help to give class labels that aligns better with the document content. Additional examples of DocKD-generated data are available in Appx.\,\ref{appx:generated_annotations}.

Table~\ref{tab:data_stats} shows some statistics of the data generated by KD and DocKD. For entity extraction, we calculate the number of unique entity types (\# of ent. types) and average number of entities generated per document (\# of ent. per doc.). We note that DocKD can generate significantly more entities and entity types than KD, by leveraging external document knowledge. Similarly, we also summarize the number unique document labels generated by KD and DocKD for document classification. For both the positive and negative class labels, DocKD generates more unique labels than KD. We attribute this to leveraging document descriptions for generation which helps LLMs generating fine-grained labels that align better with the document.

\begin{table}[!t]
    \centering
    \small
    \addtolength{\tabcolsep}{-1pt}
    \resizebox{\linewidth}{!}{
        \begin{tabular}{l|c|c|c|c}
         & \multicolumn{2}{c|}{entity extraction} & \multicolumn{2}{c}{document classification} \\
        \cline{2-5}
        method & \!\! \# of ent. types \!\! & \!\! \# ent. per doc. \!\! & \# pos. labels & \# neg. labels \\
        \Xhline{2\arrayrulewidth}
        KD    & 1454 & 11.5 & 4674 & 2476 \\
        DocKD & 2316 & 20.1 & 6053 & 3013 \\
        \end{tabular}
    }
    \vspace*{-5pt}
    \caption{Statistics of data generated by KD and DocKD.}
    \label{tab:data_stats}
\end{table}

\subsection{Leveraging Human-Labeled Annotations}
\label{subsec:analysis}

\paragraph{Human annotation QAs.}
We demonstrate that unsupervised knowledge from an LLM remains valuable even when human annotations are available for training. As shown in Table\,\ref{tab:document_vqa_with_dude}\,(a), augmenting DocVQA human annotations with DocKD-generated QAs, which incorporate a variety of document knowledge, results in stronger student models, achieving 83.4\% ANLS on the DocVQA validation set. In a more practical scenario where human-labeled documents have different distribution, we utilize DUDE, a dataset featuring multi-domain documents with diverse VQA annotations (text, numerical, yes/no, lists, etc.). In Table\,\ref{tab:document_vqa_with_dude}\,(b), DocKD-generated data significantly enhances student model performance, reaching 79.1\% ANLS, compared to 66.0\% with human annotations alone.

\begin{table}[!t]
    \centering
    \small
    \resizebox{\linewidth}{!}{
    \begin{tabular}{cc|cc|cc}
    & & \multicolumn{2}{c|}{\!\!DocVQA val\!\!} & \multicolumn{2}{c}{DUDE val} \\
    human anno. & \!\!\textbf{DocKD}-generated\!\! & ANLS & EM & ANLS & EM \\
    \Xhline{2\arrayrulewidth}
    \multicolumn{6}{c}{\textcolor{gray}{\textit{\textbf{(a)} human anno.\,=\,DocVQA train set}}} \\
    \hline
    \cmark & & 80.6 & 72.0 & 53.8 & 37.2 \\
     & \cmark & 77.2 & 68.6 & 52.6 & 36.0 \\
    \cmark & \cmark & \textbf{83.4} & \textbf{76.2} & \textbf{55.3} & \textbf{38.8} \\
    \hline
    \multicolumn{6}{c}{\textcolor{gray}{\textit{\textbf{(b)} human anno.\,=\,DUDE train set}}} \\
    \hline
    \cmark & & 66.0 & 54.9 & 54.4 & 40.0 \\
     & \cmark & 77.2 & 68.6 & 52.6 & 36.0 \\
    \cmark & \cmark & \textbf{79.1} & \textbf{70.8} & \textbf{58.0} & \textbf{42.1} \\
    \end{tabular}
    }
    \vspace*{-5pt}
    \caption{The document VQA task performance using a human-annotated training dataset. \textbf{DocKD} indicates the generated QAs from the IDL documents.
    The teacher model is Claude-2, and the student model is DFv2$_\text{base}$. For results with DFv2$_\text{large}$, refer to Appx.\,\ref{subsec:additional_results_docvqa}.
    }
    \label{tab:document_vqa_with_dude}
    \vspace*{5pt}
\end{table}

\begin{table}[!t]
    \centering
    \small
    \addtolength{\tabcolsep}{-1pt}
    \resizebox{\linewidth}{!}{
    \begin{tabular}{l|c|c|c|c|c}
     & \multicolumn{2}{c|}{RVL-CDIP test} & \multicolumn{3}{c}{out-of-domain} \\
    \cline{2-6}
    model & \!\!$\mathcal{C}_1$\,(known)\!\! & \!\!$\mathcal{C}_2$\,(unk.)\!\! & RVL-O & IRS-50 & \!\!WikiDoc\!\! \\
    \Xhline{2\arrayrulewidth}
    Falcon-40B & 62.3 & 27.4 & 76.3 & 54.0 & 39.8 \\
    \hline
    DFv2$_{\text{base}}$ S & \textbf{86.1} & 0.08 & 0.00 & 0.00 & 0.00 \\
    DFv2$_{\text{base}}$ U & 50.5 & \textbf{56.1} & 42.6 & 74.0 & 44.4 \\
    DFv2$_{\text{base}}$ S+U & 77.1 & 52.1 & \textbf{52.8} & \textbf{82.0} & \textbf{45.2} \\
    \end{tabular}
    }
    \vspace*{-5pt}
    \caption{Open-set classification performance. 
    S: supervised training with $\mathcal{C}_1$ annotations, U: unsupervised DocKD from LLM-generated class labels.}
    \label{tab:document_classification_openset}
\end{table}

\vspace*{-3pt}
\paragraph{Open-set classification.}
One of the main applications by distilling LLM's knowledge lies in its open-set classification ability, \emph{i.e.}, it can classify documents of unseen categories. The diversity of generated class labels ensures robustness, while a fixed set of annotations makes it hard to adapt to unseen labels. To verify this, let $\mathcal{C}$ denote the set of all RVL-CDIP labels, and we split $\mathcal{C}$ into two sets: $\mathcal{C}_1=$ \{{email}, {letter}, {memo}, {news article}\} and $\mathcal{C}_2=\mathcal{C}-\mathcal{C}_1$.
We train the model with documents from the web, crawled by $\mathcal{C}_1$ labels \cite{larson2022rvlcdip-o}.
Table\,\ref{tab:document_classification_openset} shows that this supervised model (S) makes highly biased predictions---while it predicts known classes accurately (86.1\%), it struggles to identify unknown categories in $\mathcal{C}_2$. In contrast, DocKD without any supervised data (U) enables generalization to unseen types of documents. Further, merging the $\mathcal{C}_1$ annotations with the generated data (S+U) leverages the advantages of both supervised and unsupervised learning.

We also evaluate our model in a more realistic distribution of data and labels, using the documents out of the domain of IDL or RVL-CDIP. To this end, we use three evaluation sets,
RVL-O \cite{larson2022rvlcdip-o}, IRS-50, and WikiDoc \cite{fujinuma2023multi}, all of which contain out-of-domain documents (refer to Appx.\,\ref{appx:dataset_specifications} for the details of datasets).
While the supervised model cannot handle these novel categories, unsupervised DocKD makes the student model even adaptable to out-of-domain classification and outlier detection, following the LLM teacher's robust predictions.

\section{Conclusion}
\label{sec:conclusion}

We address the open-world document understanding problem by instructing the LLMs to generate document annotations, given the generation prompt and OCR text. To successfully achieve this, we suggest DocKD framework, designing task prompts and answers that LLMs can easily generate, and incorporate external document knowledge from various sources. Consequently, the student models distilled by DocKD annotations demonstrate remarkable performance improvements compared to the plain KD approach in various document tasks. The integration with human-labeled annotations further enhances model performance.

\section*{Limitations}
\label{sec:limitations}

This study represents the pioneering work to utilize LLMs for open-world document understanding, specifically focusing on relatively simpler documents and tasks. We have applied LLMs to generate document annotations, and subsequently, trained student VDU models using these annotations. Our primary focus has been on common document understanding tasks such as visual question answering, entity extraction, and classification, which primarily involve documents containing tables, layouts, and forms.

However, extending our approach to handle documents with more complex visual elements, such as intricate figures, diagrams, or dense equations, remains an area for future exploration. While addressing more sophisticated problems could significantly enhance the model's applicability, such advancements would require efforts in developing new generative prompts. Furthermore, integrating LLMs with document expert models and large multimodal models, such as GPT-4V, holds potential to synthesize visually-rich, informative annotations. This integration has not yet been explored and represents a promising avenue for future research. Despite these limitations, our study lays foundational work for more complex applications in the field of document understanding using LLMs.

\bibliography{custom}

\clearpage
\appendix

\renewcommand{\contentsname}{Table of Contents for Appendix}
\renewcommand\ptctitle{} 

\part{\Large Appendix} 
\vspace*{-10pt}
\parttoc 

\section{Additional Experiments}
\label{appx:ablation_studies}

\subsection{Statistical Significance of Document Understanding Results}

We have conducted further experiments to substantiate our findings about statistical significance. Specifically, we reproduced the main results across all three tasks (Table\,\ref{tab:document_vqa}) by rerunning the experiments for the configurations DocFormerv2$_\text{large}$ + KD and DocFormerv2$_\text{large}$ + DocKD using three different random seeds. The results of these additional runs are summarized in Table\,\ref{tab:statistical_significance}. These results underscore the statistical significance and reliability of our approach.

\begin{table}[!h]
\centering
\small
\vspace{30pt}
\addtolength{\tabcolsep}{-1pt}
\resizebox{\linewidth}{!}{
\begin{tabular}{l|cc|cc|cc}
 & \multicolumn{2}{c|}{(a) VQA} & \multicolumn{2}{c|}{(b) Entity extraction} & \multicolumn{2}{c}{(c) Classification} \\
Model & val ANLS & val EM & test F1 & test ANLS & test mAcc & test mAcc* \\
\Xhline{2\arrayrulewidth}
KD run \#1 & 76.88 & 67.38 & 30.20 & 51.81 & 58.57 & 68.99 \\
KD run \#2 & 76.28 & 66.97 & 32.70 & 48.72 & 60.07 & 66.81 \\
KD run \#3 & 75.71 & 66.24 & 28.90 & 49.77 & 61.30 & 70.90 \\
KD & \!\!76.29$_{\pm 0.59}$\!\! & \!\!66.86$_{\pm 0.58}$\!\! & \!\!30.60$_{\pm 1.93}$\!\! & \!\!50.10$_{\pm 1.57}$\!\! & \!\!59.98$_{\pm 1.37}$\!\! & \!\!68.90$_{\pm 2.05}$\!\! \\
\hline
DocKD run \#1 & 81.00 & 71.85 & 61.46 & 68.66 & 62.40 & 73.93 \\
DocKD run \#2 & 80.59 & 72.16 & 62.95 & 70.29 & 63.17 & 74.76 \\
DocKD run \#3 & 80.10 & 71.60 & 62.95 & 69.58 & 63.88 & 73.93 \\
DocKD & \!\!80.56$_{\pm 0.45}$\!\! & \!\!71.87$_{\pm 0.28}$\!\! & \!\!62.45$_{\pm 0.86}$\!\! & \!\!69.51$_{\pm 0.82}$\!\! & \!\!63.15$_{\pm 0.74}$\!\! & \!\!74.21$_{\pm 0.48}$\!\! \\
\end{tabular}
}
\caption{Statistical significance of our experiments on document understanding tasks. Run \#1 are the results reported in the main paper. KD and DocKD are the results with mean $\pm$ standard deviation of the three runs.}
\label{tab:statistical_significance}
\end{table}

\subsection{Additional Results on DocVQA}
\label{subsec:additional_results_docvqa}

\paragraph{DFv2$_\text{large}$ model performance.}
Table\,\ref{tab:document_vqa_with_dude_large} presents the DocVQA validation performance with DFv2$_\text{large}$ trained on the human-annotated dataset, as in Table\,\ref{tab:document_vqa_with_dude} with DFv2$_\text{base}$. Generated QAs by DocKD are comparable to the human-labeled train set, whereas human annotations with a significantly different distribution (\emph{e.g.}, DUDE \cite{van2023dude}) may even degrade performance.

\begin{table}[!t]
    \centering
    \small
    \vspace*{5pt}
    \resizebox{.9\linewidth}{!}{
    \begin{tabular}{cc|cc}
    human anno. & \!\!\textbf{DocKD}-generated\!\! & val ANLS & val EM \\
    \Xhline{2\arrayrulewidth}
    \multicolumn{4}{c}{\textcolor{gray}{\textit{(a) human anno.\,=\,DocVQA train set}}} \\
    \hline
    \cmark & & 85.4 & 77.7 \\
     & \cmark & 81.0 & 71.9 \\
    \cmark & \cmark & \textbf{86.1} & \textbf{79.1} \\
    \hline
    \multicolumn{4}{c}{\textcolor{gray}{\textit{(b) human anno.\,=\,DUDE train set}}} \\
    \hline
    \cmark & & 74.8 & 64.0 \\
     & \cmark & \textbf{81.0} & \textbf{71.9} \\
    \cmark & \cmark & {80.3} & {71.6} \\
    \end{tabular}
    }
    \vspace*{-5pt}
    \caption{DocVQA validation performance using a human-annotated training dataset, (a) DocVQA train set and (b) DUDE train set. \textbf{DocKD} indicates the generated QAs from the IDL documents. The teacher model is Claude-2, and the student model is DFv2$_\text{large}$.
    }
    \label{tab:document_vqa_with_dude_large}
\end{table}

\paragraph{DocVQA test set performance.}
In Table\,\ref{tab:document_vqa_test}, we provide the test set performance on DocVQA  \cite{mathew2021docvqa}, in order to compare with the previous VDU models, which were all trained on the DocVQA training set.

\begin{table}[!h]
    \centering
    \footnotesize
    \addtolength{\tabcolsep}{-1.5pt}
    \resizebox{\linewidth}{!}{
    \begin{tabular}{l|l|c}
    model & size & ANLS \\
    \Xhline{2\arrayrulewidth}
    \multicolumn{3}{l}{\textcolor{gray}{\textit{DocVQA supervised learning}}} \\
    \hline
    Donut$_\text{base}$ \cite{kim2022ocr} & 143M & 67.5 \\
    T5$_\text{large}$ \cite{raffel2020t5} & 750M & 70.4 \\
    LayoutLMv2$_\text{large}$ \cite{xu2020layoutlmv2} & 426M & 86.7 \\
    LayoutLMv3$_\text{large}$ \cite{huang2022layoutlmv3} & 368M & 83.4 \\
    UDOP \cite{tang2023udop} & 794M & 84.7 \\
    DocFormerv2$_{\text{large}}$ \cite{appalaraju2023docformerv2} & 750M & 86.3$^\dagger$\!\! \\
    \hline
    \multicolumn{3}{l}{\textcolor{gray}{\textit{Training with {Claude-2}-generated data}}} \\
    \hline
    DocFormerv2$_{\text{large}}$ + KD QA & 750M & 75.8 \\
    DocFormerv2$_{\text{large}}$ + \textbf{DocKD QA} & 750M & 80.6 \\
    DocFormerv2$_{\text{large}}$ + \textbf{DocKD QA} (+\,DocVQA anno.) & 750M & \textbf{86.9} \\
    \end{tabular}
    }
    \vspace{-5pt}
    \caption{DocVQA test set performance. 
    The KD baseline uses raw OCR text for the QA generation, while DocKD uses the linearized OCR text. $^\dagger$: reproduced without searching hyperparameters. The same hyperparameters were used for training with DocKD QAs.
    }
    \label{tab:document_vqa_test}
\end{table}


\subsection{Data Volume and Quality}
\begin{figure}[!t]
    \centering
    \footnotesize
    \begin{subfigure}[t]{0.7\linewidth}
        \includegraphics[width=0.9\linewidth, left]{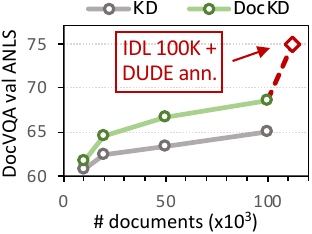}
        \caption{Falcon-40B teacher}
    \end{subfigure}%
    \vspace{10pt}
    \begin{subfigure}[t]{0.7\linewidth}
        \includegraphics[width=0.9\linewidth,left]{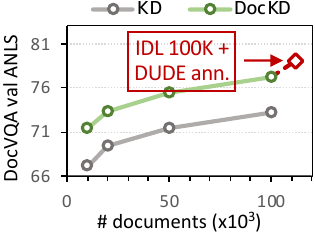}
        \caption{Claude-2 teacher}
    \end{subfigure}
    \caption{DocVQA  \cite{mathew2021docvqa} results according to the number of generated data. $x$-axis is the number of IDL  \cite{idl} documents used by the LLM to generate the QA pairs.
    }
    \label{fig:document_vqa_datasize}
    \vspace{10pt}
\end{figure}

In \fig\ref{fig:document_vqa_datasize}, we emphasize the significance of the distilled data volume in capturing diverse knowledge. Additionally, the introduction of a small set of human annotations (\emph{e.g.}, DUDE \cite{van2023dude}) from a different domain proves beneficial, especially when the teacher model size is small and thus generates data of lower quality.

However, it is crucial to note that a larger volume of generated data does not always guarantee superior performance, \emph{i.e.}, quality of the dataset is also important. For the classification task, we established evaluation criteria for generated labels, accounting for both word length and frequency within the dataset. Labels exceeding a word length of 5 (considered overly specific) or occurring less than 3 times throughout the dataset (outliers) were excluded. Documents without remaining positive labels were removed, consequently reducing our IDL training set size from 50K to 43K. This refinement enhanced overall data quality, resulting in an improved test accuracy (+3.5\%). Similarly, in VQA and entity extraction tasks, we filtered out excessively long or short questions/answers and field names identified as outliers.

\subsection{Using Human-Labeled FUNSD Entities}
For the entity extraction task, we utilized RVL-CDIP invoices \cite{harley2015rvlcdip}, extracting keys and values, and applying the entity generation prompts. Here, we use FUNSD \cite{jaume2019funsd} dataset, which is a small subset of RVL-CDIP forms, and all the KV entities are manually annotated. In this case, we use their annotations for the KV entity inputs.
Table\,\ref{tab:entity_extraction_with_funsd} shows that,
although FUNSD contains only a small number of document samples, an LLM can generate reliable KV entity fields based on the manual annotations. Combining with invoices documents that have abundant entities, the student model is effectively distilled with diverse knowledge and can exhibit the highest entity extraction performances.

\begin{table}[!t]
    \centering
    \small
    \addtolength{\tabcolsep}{-1pt}
    \resizebox{\linewidth}{!}{
    \begin{tabular}{l|l|l|c|c}
    teacher & gen. data (\# doc.) & \# entities & CORD & DeepForm \\
    \Xhline{2\arrayrulewidth}
    Falcon-40B & FUNSD (149) & 2,308 & 33.2 & 44.6 \\
    Falcon-40B & Invoices (5,000) & 38,121 & \textbf{55.1} & 48.5 \\
    Falcon-40B & FUNSD\,+\,Invoices & 40,429 & 54.9 & \textbf{52.2} \\
    \hline
    Claude-2 & FUNSD (149) & 2,608 & 42.8 & 49.1 \\
    Claude-2 & Invoices (5,000) & 74,289 & 60.2 & 64.2 \\
    Claude-2 & FUNSD\,+\,Invoices & 76,897 & \textbf{60.4} & \textbf{67.5} \\
    \end{tabular}
    }
    \vspace*{-5pt}
    \caption{Entity extraction from FUNSD \cite{jaume2019funsd} and RVL-CDIP invoices \cite{harley2015rvlcdip} documents. The student model is DFv2$_\text{base}$.}
    \label{tab:entity_extraction_with_funsd}
\end{table}

\subsection{Ablation Study on Entity Generation Strategies}

In the entity extraction task, we have utilized the LLM's entity recognition ability and the KV detection model's key-value extraction ability. To unveil the individual contributions of each component, Table\,\ref{tab:ner_ablation} presents an ablation study on different entity generation methods. Using only $\mathbf{p}_{\text{gen-}\textit{ent}}$ represents the plain KD baseline without external document knowledge. On the other hand, using only $\mathbf{p}_{\text{gen-}\textit{kv}}$ eliminates the LLM's automatic extraction of entities that are not detected as keys or values. In addition to these approaches, we conduct key normalization method, where the LLM generates variants for each key name, and these normalized variants serve as the field for the KV entities. This method does not utilize KV entity constraints, which have been used in DocKD as an iterative presentation of KV entities for consistency with previous entities and fields.

The ablation study results confirm the significace of both $\mathbf{p}_{\text{gen-}\textit{ent}}$ and $\mathbf{p}_{\text{gen-}\textit{kv}}$, coupled with KV detection. Notably, providing the LLM with detected KV pairs yields substantial improvement ($\mathbf{p}_{\text{gen-}\textit{ent}}$ vs. DocKD), while the extraction of non-KV entities also proves to be crucial ($\mathbf{p}_{\text{gen-}\textit{kv}}$ vs. DocKD). Injecting context on previous KV entities and the generated fields further enhances the reliability of subsequent generation (key normalization vs. DocKD).

\begin{table}[!t]
    \centering
    \small
    \addtolength{\tabcolsep}{-2pt}
    \resizebox{\linewidth}{!}{
    \begin{tabular}{l|ccc|c}
    method & \makecell[b]{Entity\\recognition} & \makecell[b]{KV\\detection} & \makecell[b]{KV\\constraints} & F1 \\
    \Xhline{2\arrayrulewidth}
    $\mathbf{p}_{\text{gen-}\textit{ent}}$ (KD) & \cmark & \xmark & \xmark & 20.9 \\
    key normalization & \xmark & \cmark & \xmark & 39.2 \\
    $\mathbf{p}_{\text{gen-}\textit{kv}}$ & \xmark & \cmark & \cmark & 45.6 \\
    \hline
    $\mathbf{p}_{\text{gen-}\textit{ent}} + \mathbf{p}_{\text{gen-}\textit{kv}}$ (\textbf{DocKD}) & \cmark & \cmark & \cmark & \textbf{55.1}
    \end{tabular}
    }
    \caption{Ablation study on CORD \cite{park2019cord} entity extraction. Entities and field names are generated from 5K RVL-CDIP invoices \cite{harley2015rvlcdip} by the Falcon-40B \cite{almazrouei@falcon40b} teacher. The student model is DFv2$_\text{base}$. Note that $\mathbf{p}_{\text{gen-}\textit{kv}}$ always requires the KV detection in prior.}
    \label{tab:ner_ablation}
\end{table}

\subsection{Ablation Study on the Effect of Descriptions}
In the document classification task, descriptions play a crucial role in two key aspects: generating positive labels and appending descriptions when constructing the candidate list. To assess the effect of each aspect, we establish an ablation baseline, KD L+D, and compare three distillation methods:
\begin{itemize}[label={$\circ$}, leftmargin=*]
    \item \textbf{KD L}: LLM generates only class labels without any description.
    \item \textbf{KD L+D}: LLM generates description and, in sequence, class labels based on the description. However, it does not append the desciptions to the class labels during the formulation of the candidate list. 
    \item \textbf{DocKD L+D}: LLM generates description and, in sequence, class labels based on the description. These descriptions are appended to the candidate list to give a hint about the class.
\end{itemize}
Table\,\ref{tab:desc_ablation} substantiates the efficacy of utilizing descriptions in both aspects. However, the superior performance gain is observed when appending descriptions to the candidate list. This suggests that designing the task prompt to incorporate rich information about the labels is an effective strategy in training the student model.

\begin{table}[!t]
    \centering
    \small
    \begin{tabular}{l|c|c}
    method & mAcc & mAcc$^\star$ \\
    \Xhline{2\arrayrulewidth}
    KD L & 56.3 & 63.4 \\
    KD L+D & 57.9 & 68.4 \\
    \textbf{DocKD L+D} & \textbf{61.9} & \textbf{74.0} \\
    \end{tabular}
    \caption{Ablation study on RVL-CDIP \cite{harley2015rvlcdip} classification. The student model is DFv2$_\text{base}$, and the teacher model is Claude-2.}
    \label{tab:desc_ablation}
\end{table}

\subsection{Full Results of RVL-CDIP Classification}
\label{appx:classification_full_results}

Table\,\ref{tab:rvl_cdip_full} shows the full category results for document classification, which were sumarized into mean accuracy in Table\,\ref{tab:document_vqa}\,(c).

\begin{table*}[!t]
    \centering
    \small
    \addtolength{\tabcolsep}{-2pt}
    \resizebox{\textwidth}{!}{
    \renewcommand{\arraystretch}{1.1}
    \begin{tabular}{lccccccccccccccccc}
    model & \rotatebox[origin=l]{90}{letter} & \rotatebox[origin=l]{90}{form} & \rotatebox[origin=l]{90}{email} & \rotatebox[origin=l]{90}{handwritten} & \rotatebox[origin=l]{90}{advertisement} & \rotatebox[origin=l]{90}{scientific report} & \rotatebox[origin=l]{90}{scientific publication} & \rotatebox[origin=l]{90}{specification} & \rotatebox[origin=l]{90}{file folder} & \rotatebox[origin=l]{90}{news article} & \rotatebox[origin=l]{90}{budget} & \rotatebox[origin=l]{90}{invoice} & \rotatebox[origin=l]{90}{presentation} & \rotatebox[origin=l]{90}{questionnaire} & \rotatebox[origin=l]{90}{resume} & \rotatebox[origin=l]{90}{memo} & mAcc \\
    \Xhline{2\arrayrulewidth}
    \multicolumn{10}{l}{\textcolor{gray}{\textit{LLM zero-shot prediction}}} \\
    \hline
    Flan-T5$_\text{large}$\,\cite{chung2022flan} & 15.0 & 8.2 & 66.5 & 0.3 & 68.3 & 50.2 & 91.0 & 62.5 & 4.2 & 59.9 & 29.6 & 83.7 & 19.9 & 62.5 & 50.1 & 73.0 & 46.6 \\
    Flan-T5$_\text{XXL}$\,\cite{chung2022flan} & 36.5 & 31.7 & 88.8 & 5.0 & 65.0 & 50.8 & 44.2 & 58.7 & 11.3 & 80.4 & 26.7 & 75.4 & 32.5 & 77.5 & 61.6 & 86.4 & 52.0 \\
    LLaVA-1.5\,\cite{liu2023improved} & 88.2 & 53.8 & 7.5 & 21.3 & 72.5 & 45.3 & 22.3 & 35.4 & 6.7 & 60.0 & 40.8 & 69.6 & 3.8 & 6.4 & 17.9 & 26.9 & 36.1 \\
    Vicuna-1.3\,\cite{vicuna2023} & 62.3 & 30.4 & 87.8 & 1.7 & 68.5 & 84.6 & 67.4 & 76.7 & 0.2 & 73.1 & 28.3 & 60.5 & 21.9 & 52.0 & 0.9 & 57.9 & 48.4 \\
    Falcon\,\cite{almazrouei@falcon40b} & 67.3 & 14.8 & 65.7 & 10.2 & 50.3 & 59.0 & 18.4 & 49.5 & 4.9 & 66.9 & 10.5 & 55.7 & 11.5 & 39.2 & 21.9 & 60.7 & 37.9 \\
    \hline
    \multicolumn{10}{l}{\textcolor{gray}{\textit{VDU models trained with \textcolor{Tan}{\textbf{only generated}} data}}} \\
    \hline
    Flan-T5$_\text{large}$ + KD & 36.6 & 23.0 & 21.7 & 2.3 & 89.5 & 64.5 & 90.6 & 76.1 & 20.7 & 61.4 & 31.4 & 68.7 & 34.8 & 74.4 & 79.2 & 61.5 & 52.3 \\
    Flan-T5$_\text{large}$ + \textbf{DocKD} & 72.6 & 9.1 & 89.7 & 3.2 & 86.4 & 68.9 & 77.2 & 73.9 & 5.1 & 76.1 & 40.4 & 84.4 & 29.8 & 85.3 & 96.7 & 12.4 & 57.0 \\
    DocFormerv2$_\text{large}$ + KD & 59.3 & 17.5 & 75.2 & 0.9 & 91.5 & 69.9 & 87.4 & 76.2 & 22.2 & 67.9 & 29.3 & 73.5 & 38.5 & 85.7 & 94.6 & 47.7 & 58.6 \\
    DocFormerv2$_\text{large}$ + \textbf{DocKD} & 55.8 & 21.4 & 89.6 & 6.7 & 78.2 & 55.5 & 89.8 & 87.4 & 6.6 & 85.4 & 56.1 & 79.4 & 26.3 & 92.2 & 96.3 & 71.8 & \textbf{62.4} \\
    \Xhline{2\arrayrulewidth}
    \end{tabular}
    }
    \caption{RVL-CDIP classification results of all 16 categories.}
    \label{tab:rvl_cdip_full}
\end{table*}
\renewcommand{\ttfamily}{\fontencoding{T1}\fontfamily{lmtt}\selectfont}

\section{Generation Prompts for LLMs}
\label{appx:generative_prompts}

We provide full templates for the generation prompts $\mathbf{p}_\text{gen}$, which are input to the LLM in conjunction with the document text. The generation prompts enable the LLM to proficiently generate document annotations, which are further used to train student models.

\subsection{Generation Prompt for Document VQA}

In the document VQA task, the generation prompt serves as a guidance for the LLM to generate a fixed number of question-answer (QA) pairs, which can be answered by referencing the document's OCR text. To facilitate this process, we provide two instructive examples and articulate several rules. Then, for the specific target document, which is an IDL \cite{idl} document in our study, we extract OCR text from the image, convert it to linearized text (refer to Sec.\,\ref{subsec:vqa}), and embed this text into the placeholder \texttt{\{LINEARIZED\_TEXT\_PLACE\_HOLDER\}} in $\mathbf{p}_\text{gen}$. We set \texttt{\{COUNT\_PLACE\_HOLDER\}} to three.

\vspace*{20pt}
\noindent
\begin{tcolorbox}[space to upper,
    skin=bicolor,
    boxsep=1.5pt,
    colbacklower=black!80,
    collower=white,
    colframe=black!85,
    title={\scriptsize $\mathbf{p}_\text{gen}$ for QA pair generation},
    halign=left,
    valign=top,
    nobeforeafter,
    halign lower=flush right,
    bottom=2mm,
    height=14.2cm,
    breakable
    ]
    \scriptsize
    \texttt{{[}Example 1{]} \\
    \, \\
    Document: Confidential RJRT PR APPROVAL DATE: 1/8/93 SUBJECT: Ru IVAs PROPOSED RELEASE DATE: for response FOR RELEASE TO: CONTACT: P. CARTER ROUTE TO: Name Initials Date Peggy Carter Ace 1/1/15 Kaura Payne nt. T/R Return to Peggy Carter, PR, 16 Reynolds Building Not \\
    \, \\
    Generate three question-answer pairs from this document. \\
    \, \\
    Question: what is the date mentioned in this letter? \\
    Answer: 1/8/93 \\
    \, \\
    Question: what is the contact person name mentioned in this letter? \\
    Answer: P. Carter \\
    \, \\
    Question: What is the address of Peggy Carter? \\
    Answer: 16 Reynolds Building \\
    \, \\
    \, \\
    {[}Example 2{]} \\
    \, \\
    Document: Link between IR and CVD THE ROUTE TO CARDIOVASCULAR DISEASE 2.11.15-19 Hyperglycemia Insulin Hyper a path that leads to increased risk for MI Resistance Dys TYPE 2 DIABETES EQUALS PRIOR MI AS A CHD RISK FACTOR Pr S 7-year incidence of myocardial infarction (MI) (\%) 25\% 20\% 15\% 18.8\% 20.2\% 10\% 5\% 0\% Nondiabetic patients Type 2 diabetics with prior MI without prior MI \\
    \, \\
    Generate two question-answer pairs from this document. \\
    \, \\
    Question: Heading of the document? \\
    Answer: Link between IR and CVD \\
    \,\\
    Question: what does MI stand for? \\
    Answer: myocardial infarction \\
    \,\\
    \,\\
    Rules: \\
    - Use the following test document as the only source of information. \\
    - Make questions diverse as possible. \\
    - Answers should be simple and specified in the document. \\
    - Generate ONLY questions and answers, do not give any explanations. \\
    \,\\
    {[}Test{]} \\
    \,\\
    Document: \{LINEARIZED\_TEXT\_PLACE\_HOLDER\} \\
    \,\\
    Generate \{COUNT\_PLACE\_HOLDER\} question-answer pairs from this document.
}%
 \end{tcolorbox}%

\vspace*{10pt}
\subsection{Generation Prompt for Entity Extraction}
\label{appx:generation_prompt_entity_extraction}

We separate the generation of entities and field names into two parts: for non-KV entities and for KV entities. For the former, the generation prompt $\mathbf{p}_{\text{gen-}\textit{ent}}$ is employed to extract entities from the document text as well as assigning their names. This process is exemplified through two instructive examples. Provided with the document text, the LLM is instructed to extract entities enclosed with $<$\texttt{regular}$>$ and $<$\texttt{/regular}$>$ tags. Also, each line of entity is delimited by a separator ``\texttt{ --- }'', followed by the corresponding generated field name. Note that, to avoid duplicated generations for KV entities, we remove all the detected KV entities from the document text: \texttt{\{TEXT\_WITHOUT\_KV\_PLACE\_HOLDER\}} (refer to Sec.\,\ref{subsec:entity_extraction}).

For the KV entities identified by a KV detection model, $\mathbf{p}_{\text{gen-}\textit{kv}}$ instructs the LLM to generate only the field names for these entities. In the OCR text, the KV entities are enclosed by the tags $<$\texttt{kv}$>$ and $<$\texttt{/kv}$>$ to provide explicit guidance to the model regarding which part it should refer to. The iterative presentation of each KV entity, line by line, involves inputting each line into \texttt{\{CONSTRAINTS\_PLACE\_HOLDER\}} in the format of ``\texttt{$<$kv$>$key value$<$/kv$>$ --- }''. The generated field name is then appended to the constraint for the next iteration.

\vspace*{40pt}
\noindent
\begin{tcolorbox}[space to upper,
    skin=bicolor,
    boxsep=1.5pt,
    colbacklower=black!80,
    collower=white,
    colframe=black!85,
    title={\scriptsize $\mathbf{p}_{\text{gen-}\textit{ent}}$ for entity generation},
    halign=left,
    valign=top,
    nobeforeafter,
    halign lower=flush right,
    bottom=2mm,
    height=10.5cm,
    breakable
    ]
    \scriptsize
    \texttt{Task: I want to get entities and their entity types from OCR text of documents. \\
    \,\\
    OCR text1: Invoice us EK Packaging Goras Ice Cream \$ Kathwada GIDC EK Packaging Ahmedabad, Gujarat. \\
    \,\\
    $<$regular entities for OCR text1$>$ \\
    1. $<$regular$>$EK Packaging$<$/regular$>$ --- Company Name \\
    2. $<$regular$>$Goras Ice Cream$<$/regular$>$ --- Customer Name \\
    3. $<$regular$>$Kathwada GIDC$<$/regular$>$ --- Customer Address \\
    4. $<$regular$>$EK Packaging Ahmedabad, Gujarat.$<$/regular$>$ --- Company Address \\
    \,\\
    \,\\
    OCR text2: 1 REAL GANACHE 16,500 1 egg tart 13,000 1 pizza toast 16,000 \\
    \,\\
    $<$regular entities for OCR text2$>$ \\
    1. $<$regular$>$REAL GANACHE$<$/regular$>$ --- Item Name \\
    2. $<$regular$>$16,500$<$/regular$>$ --- Item Price \\ 
    3. $<$regular$>$egg tart$<$/regular$>$ --- Item Name \\
    4. $<$regular$>$13,000$<$/regular$>$ --- Item Price \\
    5. $<$regular$>$pizza toast$<$/regular$>$ --- Item Name \\
    6. $<$regular$>$16,000$<$/regular$>$ --- Item Price \\
    7. $<$regular$>$1$<$/regular$>$ --- Item Quantity \\
    \,\\
    \,\\
    OCR text3: \{TEXT\_WITHOUT\_KV\_PLACE\_HOLDER\} \\
    \,\\
    $<$regular entities for OCR text3$>$ \\
    1. $<$regular$>$
    }%
 \end{tcolorbox}%

\vspace*{40pt}
\noindent
\begin{tcolorbox}[space to upper,
    skin=bicolor,
    boxsep=1.5pt,
    colbacklower=black!80,
    collower=white,
    colframe=black!85,
    title={\scriptsize $\mathbf{p}_{\text{gen-}\textit{kv}}$ for KV entity generation},
    halign=left,
    valign=top,
    nobeforeafter,
    halign lower=flush right,
    bottom=0mm,
    height=9cm,
    breakable
    ]
    \scriptsize
    \texttt{Task: I want to get entities and their entity types from OCR text of documents. \\
    \,\\
    OCR text1: Invoice us EK Packaging Goras Ice Cream \$ Kathwada GIDC $<$kv$>$Inv. date 14-03-20$<$/kv$>$ EK Packaging Ahmedabad, Gujarat. $<$kv$>$Due 29-03-20$<$/kv$>$ $<$kv$>$Inv. \# 1248$<$/kv$>$ \\
    \,\\
    $<$kv entities for OCR text1$>$ \\
    1. $<$kv$>$Inv. date 14-03-20$<$/kv$>$ --- Invoice Date \\
    2. $<$kv$>$Due 29-03-20$<$/kv$>$ --- Due Date \\
    3. $<$kv$>$Inv. \# 1248$<$/kv$>$ --- Invoice Number \\
    \,\\
    \,\\
    OCR text2: 1 REAL GANACHE 16,500 1 egg tart 13,000 1 pizza toast 16,000 $<$kv$>$TOTAL 45,500$<$/kv$>$ $<$kv$>$CASH 50,000$<$/kv$>$ $<$kv$>$CHANGE 4,500$<$/kv$>$ \\
    \,\\
    $<$kv entities for OCR text2$>$ \\
    1. $<$kv$>$TOTAL 45,500$<$/kv$>$ --- Total Amount \\
    2. $<$kv$>$CASH 50,000$<$/kv$>$ --- Payment Amount \\ 
    3. $<$kv$>$CHANGE 4,500$<$/kv$>$ --- Change \\
    \,\\
    \,\\
    OCR text3: \{TEXT\_WITH\_KV\_TAGS\_PLACE\_HOLDER\} \\
    \,\\
    $<$kv entities for OCR text3$>$ \\
    \{CONSTRAINTS\_PLACE\_HOLDER\} \\
    }%
 \end{tcolorbox}%

\subsection{Generation and Inference Prompts for Document Classification}
\label{appx:generative_prompts_for_cls}

In the document classification task, we need three distinct generation prompts designed for generating descriptions, positive labels list, and negative labels list, respectively. Initially, $\mathbf{p}_{\text{gen-}\textit{desc}}$ prompts the LLM to generate a description by characterizing the document type based on the document text. Subsequently, the generated output $\mathbf{a}_{\text{gen-}\textit{desc}}$ is incorporated into the following prompt, $\mathbf{p}_{\text{gen-}\textit{pos}}$, specifically within the placeholder \texttt{\{DESCRIPTION\_PLACE\_HOLDER\}}. This serves the purpose of providing contextual information about the document, thereby facilitating the accurate generation of positive labels. Finally, the output $\mathbf{a}_{\text{gen-}\textit{pos}}$  is introduced to \texttt{\{POSITIVES\_PLACE\_HOLDER\}} in the negative generation prompt $\mathbf{p}_{\text{gen-}\textit{neg}}$. This instructs the LLM to avoid suggesting types similar to those in the positives list.

\noindent
\begin{tcolorbox}[space to upper,
    skin=bicolor,
    boxsep=1.5pt,
    colbacklower=black!80,
    collower=white,
    colframe=black!85,
    title={\scriptsize $\mathbf{p}_{\text{gen-}\textit{desc}}$ for document description generation},
    halign=left,
    valign=top,
    nobeforeafter,
    halign lower=flush right,
    bottom=3mm,
    height=2.5cm
    ]
    \scriptsize
    \texttt{Document: \{TEXT\_PLACE\_HOLDER\} \\
    \,\\
    Question: Can you describe the document type of the above document in one sentence? \\
    \,\\
    Answer:
    }%
 \end{tcolorbox}%

\vspace*{20pt}
\noindent
\begin{tcolorbox}[space to upper,
    skin=bicolor,
    boxsep=1.5pt,
    colbacklower=black!80,
    collower=white,
    colframe=black!85,
    title={\scriptsize $\mathbf{p}_{\text{gen-}\textit{pos}}$ for positive label generation},
    halign=left,
    valign=top,
    nobeforeafter,
    halign lower=flush right,
    bottom=0mm,
    height=4.2cm
    ]
    \scriptsize
    \texttt{Text of the document: \{TEXT\_PLACE\_HOLDER\} \\
    \,\\
    Short description of the document: \{DESCRIPTION\_PLACE\_HOLDER\} \\ 
    \,\\
    Question: Given the above text of a document and its short description, can you suggest a list of \{COUNT\_PLACE\_HOLDER\} possible types (or names) of the document? Please list only types, without any explanation or description. \\
    \,\\
    Answer:
    }%
 \end{tcolorbox}%

\vspace*{20pt}
\noindent
\begin{tcolorbox}[space to upper,
    skin=bicolor,
    boxsep=1.5pt,
    colbacklower=black!80,
    collower=white,
    colframe=black!85,
    title={\scriptsize $\mathbf{p}_{\text{gen-}\textit{neg}}$ for negative label generation},
    halign=left,
    valign=top,
    nobeforeafter,
    halign lower=flush right,
    bottom=0mm,
    height=4cm
    ]
    \scriptsize
    \texttt{Document: \{TEXT\_PLACE\_HOLDER\} \\
    \,\\
    Matching types list: \{POSITIVES\_PLACE\_HOLDER\} \\
    \,\\
    Question: Given the above text extracted from a document using OCR, can you suggest a list of \{COUNT\_PLACE\_HOLDER\} possible document types (or names) that do NOT match the document? Do not include types similar to the matching list. \\
    \,\\
    Answer:
    }%
 \end{tcolorbox}%

\vspace*{10pt}

\begin{figure}[!h]
    \centering
    \small
    \includegraphics[width=.75\linewidth]{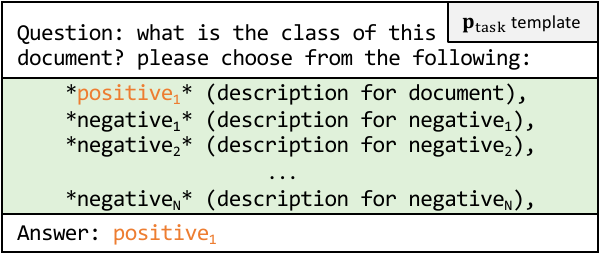}
    \vspace{-7pt}
    \caption{Classification task prompt template. The candidate list is composed of one positive label and a few negative labels, appended with descriptions.}
    \label{fig:class_generation}
    \vspace*{-15pt}
  \end{figure}

For inference, we support open-world classification by dynamically constructing a candidate list in the prompt. We ask the model to select the class label that matches best with given document. \fig\ref{fig:class_generation} shows the prompt $\mathbf{p}_\text{task}$ we used in our experiment.

\subsection{Connectivity Between the Proposed Methods}
In this study, tailoring generation prompts and document text formats for specific tasks has been proposed, and there is a potential for synergy when combining these approaches. However, the effectiveness of such combination depends on the chosen document knowledge injection method and the nature of the task. For instance, we observed that text linearization did not enhance classification accuracy and could not be transferred to entity extraction, as the field name generation also involves distinct modifications to $\mathbf{d}_\text{text}$ (refer to Appx.\,\ref{appx:generation_prompt_entity_extraction}). On the other hand, leveraging document descriptions or reasoning steps may hold promise for improving the QA generation. Yet, this would require non-trivial efforts in designing new generative prompts, and it is identified as a prospective direction for future research.

\subsection{Improving the Instructions for LLM Zero-Shot Prediction}
\label{appx:llm_inst_tuning}

While numerous strategies exist for enhancing LLM zero-shot predictions through instruction modulation, the optimal approach varies depending on the model type. Although we have not explored optimal instruction strategies for every language model, our work involves minimal engineering efforts to identify the LLM's performance in document understanding tasks and show that small student models trained by DocKD are as effective as the LLMs.

In this section, we describe our enhancements to the prompt for improving zero-shot predictions of Claude-2 and Falcon-40B models, in document VQA and classification tasks. Essentially, we provide the LLM with $\mathbf{p}_\text{task}$ and $\mathbf{d}_\text{text}$ as inputs, employing the same design as utilized for the student models. Within $\mathbf{p}_\text{task}$, we input instructions to regulate the output format for each LLM, facilitating the parsing of the answer into the desired format.

\paragraph{Instructions for DocVQA.}
We leverage linearized OCR text, a method previously employed in generating QA pairs from the LLM. Given the LLM's ability in comprehending linearized text, we convert the OCR text into the linearized form and ask the document question.
In addition, since DocVQA is an extractive QA dataset, \ie, answers are directly extracted from the provided context, we use the dataset-specific prompt to control the outputs. To achieve this, we implement instructing rules as suggested in  \cite{wang2023latin}.
This strategy has significantly increased DocVQA val ANLS to 58.3 $\rightarrow$ 79.6 for Claude-2, and 52.6 $\rightarrow$ 72.4 for Falcon-40B.
In summary, the task prompt for DocVQA is provided as follows. 

\vspace*{20pt}
\noindent
\begin{tcolorbox}[space to upper,
    skin=bicolor,
    boxsep=1.5pt,
    colbacklower=black!80,
    collower=white,
    colframe=black!85,
    title={\scriptsize $\mathbf{p}_\text{task}$ for DocVQA zero-shot prediction},
    halign=left,
    valign=top,
    nobeforeafter,
    halign lower=flush right,
    bottom=3mm,
    height=8.1cm,
    breakable
    ]
    \scriptsize
    \texttt{You are asked to answer the question based on the given document OCR text. \\
    \, \\
    For example, \\
    Context: Confidential RJRT PR APPROVAL DATE: 1/8/93 SUBJECT: Ru IVAs PROPOSED RELEASE DATE: for response FOR RELEASE TO: CONTACT: P. CARTER ROUTE TO: Name Initials Date Peggy Carter Ace 1/1/15 Kaura Payne nt. T/R Return to Peggy Carter, PR, 16 Reynolds Building Not \\
    Answer the question: What is the contact person name mentioned in this letter? \\
    Answer: P. Carter \\
    \,\\
    Rules: \\
    - The answers to questions are short text spans taken verbatim from the document. This means that the answers comprise a set of contiguous text tokens present in the document. \\
    - Directly extract the answer of the question from the document with as few words as possible. \\
    \,\\
    \,\\
    Context: \{LINEARIZED\_TEXT\_PLACE\_HOLDER\} \\
    Answer the question: \{QUESTION\_PLACE\_HOLDER\} \\
    Answer:
}%
 \end{tcolorbox}%

\paragraph{Instructions for RVL-CDIP.}
Recognizing the significance of document descriptions in enhancing knowledge utilization and improving class label generation, we adopt a 2-step classification approach. In the initial step, the LLM does not classify directly but instead generates the possible document type according to its own interpretation. Subsequently, in the second step, we provide the output from the first step into \texttt{\{TYPE\_PLACE\_HOLDER\}} as a suggested document name, and instruct the model to select the document type from the candidate list.
In addition, we recognize that Falcon-40B struggles in accurately naming the exact category, even when provided with a list. To address this, we emphasize all 16 evaluation categories.
This strategic modulation has improved RVL-CDIP test mAcc to 31.8 $\rightarrow$ 37.9 for Falcon-40B, compared to direct classification. However, Claude-2 does not achieve further performance gain through this instruction. Additionally, attempts to replace the document text with linearized text, as done in DocVQA, do not yield improvements in this task.

\vspace*{20pt}
\noindent
\begin{tcolorbox}[space to upper,
    skin=bicolor,
    boxsep=1.5pt,
    colbacklower=black!80,
    collower=white,
    colframe=black!85,
    title={\scriptsize $\mathbf{p}_\text{task}$ for RVL-CDIP zero-shot prediction},
    halign=left,
    valign=top,
    nobeforeafter,
    halign lower=flush right,
    bottom=3mm,
    height=9.3cm,
    breakable
    ]
    \scriptsize
    \texttt{Choose the document type based on the given context. We have 16 categories. \\
    \, \\
    - letter \\
    - form \\
    - email \\
    - handwritten \\
    - advertisement \\
    - scientific report \\
    - scientific publication \\
    - specification \\
    - file folder \\
    - news article \\
    - budget \\
    - invoice \\
    - presentation \\
    - questionnaire \\
    - resume \\
    - memo \\
    \,\\
    \,\\
    Context: \{TEXT\_PLACE\_HOLDER\} \\
    Suggested document name: \{TYPE\_PLACE\_HOLDER\} \\
    Question: What is the document type of this document? Please choose from the following: \{letter; form; email; handwritten; advertisement; scientific report; scientific publication; specification; file folder; news article; budget; invoice; presentation; questionnaire; resume; memo\} \\
    Answer: 
}%
 \end{tcolorbox}%

\section{Examples of Generated Annotations}
\label{appx:generated_annotations}

We present the examples of LLM-generated annotations, for document VQA in Appx.\,\ref{subsec:generated_qas}, for entity extraction in Appx.\,\ref{subsec:generated_entities}, and for document classification in Appx.\,\ref{subsec:generated_classes}.

\subsection{Generated QAs for Document VQA}
\label{subsec:generated_qas}

\paragraph{Using raw OCR text vs. linearized OCR text.}
Table\,\ref{tab:generated_qas-1} and Table\,\ref{tab:generated_qas-2} describe the generated QAs from Claude-2, comparing the results from the plain KD (using raw OCR text) and DocKD (using linearized OCR text). In Table\,\ref{tab:generated_qas-1}, the document includes line numbers for each line of text, but raw OCR text lacks this structural detail, resulting in misplaced numbers in the middle of text. Consequently, Claude-2 generates inaccurate questions, such as \texttt{Question 1} erroneously referencing a non-existent question number 2, or \texttt{Question 2} inquiring about the percentage of children, which cannot be directly answered from the document.
In contrast, when linearized OCR text is utilized, questions align with the document context, ensuring correct answers. Notably, questions explicitly refer to line numbers, \eg, inquiring about the contents in \texttt{line 1} or in \texttt{lines 5--8}, which requires visual knowledge to answer.

In Table\,\ref{tab:generated_qas-2}, the document contains words and numbers in a structured form, posing a challenge for the LLM in generating informative QAs from the OCR text. In KD QAs, \texttt{Question 1} and \texttt{Question 3} are easily extracted and straightforward to answer without visual knowledge. \texttt{Question 2}, which pertains to tabular information, is paired with \texttt{Answer 2}, which is incorrect. In contrast, \texttt{Question 2} of DocKD requires reference to the table format, specifically in the third row and the second column, for a correct response. Also, the paired \texttt{Answer 2} is correct. Similarly, \texttt{Question 3} and \texttt{Answer 3} are about the contents in the second row and the last column of the table.

\begin{figure*}[!ht]
    \centering
    \includegraphics[width=.8\linewidth]{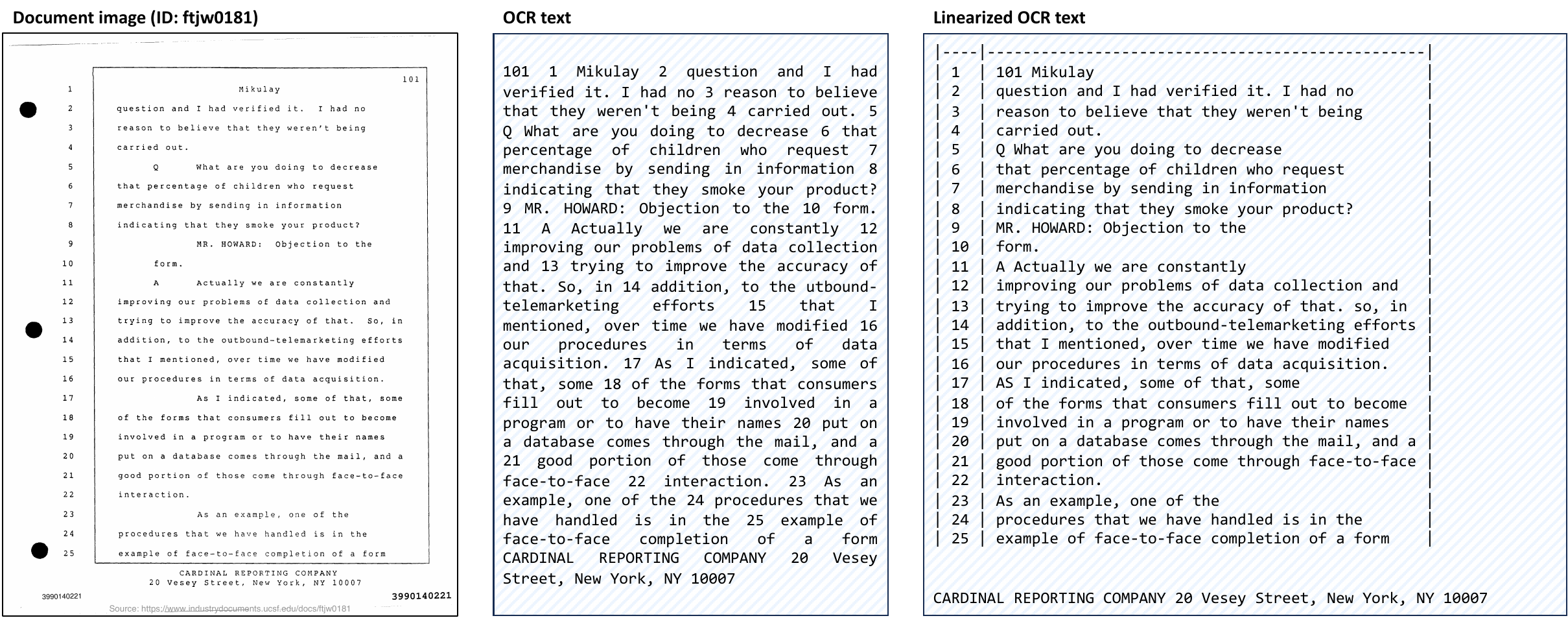}
\end{figure*}
\begin{table*}[!ht]
    \centering
    \scriptsize
    \begin{tabular*}{.8\linewidth}{l}
    \toprule
    KD QAs (using OCR text)\\
    \midrule
    \ttfamily
    Question 1: What is the question number 2 referring to in the document? \\
    \ttfamily
    Answer 1: Mikulay \\
    \,\\
    \ttfamily
    Question 2: What percentage of children request merchandise by sending in information \\
    \ttfamily
    indicating that they smoke the company's product? \\
    \ttfamily
    Answer 2: 101 \\
    \,\\
    \ttfamily
    Question 3: What are some of the ways the company collects data mentioned in the document? \\
    \ttfamily
    Answer 3: Through outbound telemarketing efforts, mail, and face-to-face interaction. \\
    \bottomrule
    \toprule
    \textbf{DocKD QAs} (using linearized OCR text) \\
    \midrule
    \ttfamily
    Question 1: What is the name mentioned in line 1? \\
    \ttfamily
    Answer 1: Mikulay \\
    \,\\
    \ttfamily
    Question 2: What does the questioner ask about in lines 5-8? \\
    \ttfamily
    Answer 2: What the person is doing to decrease the percentage of children who request \\
    \ttfamily
    merchandise by sending in information indicating that they smoke the product. \\
    \,\\
    \ttfamily
    Question 3: What does the respondent say they are constantly improving in lines 11-12? \\
    \ttfamily
    Answer 3: Their problems of data collection. \\
    \bottomrule
    \end{tabular*}
    \vspace{-5pt}
    \caption{Generated QA pairs for the IDL document (ID: ftjw0181).}
    \label{tab:generated_qas-1}
\end{table*}

\begin{figure*}[!ht]
    \centering
    \includegraphics[width=.8\linewidth]{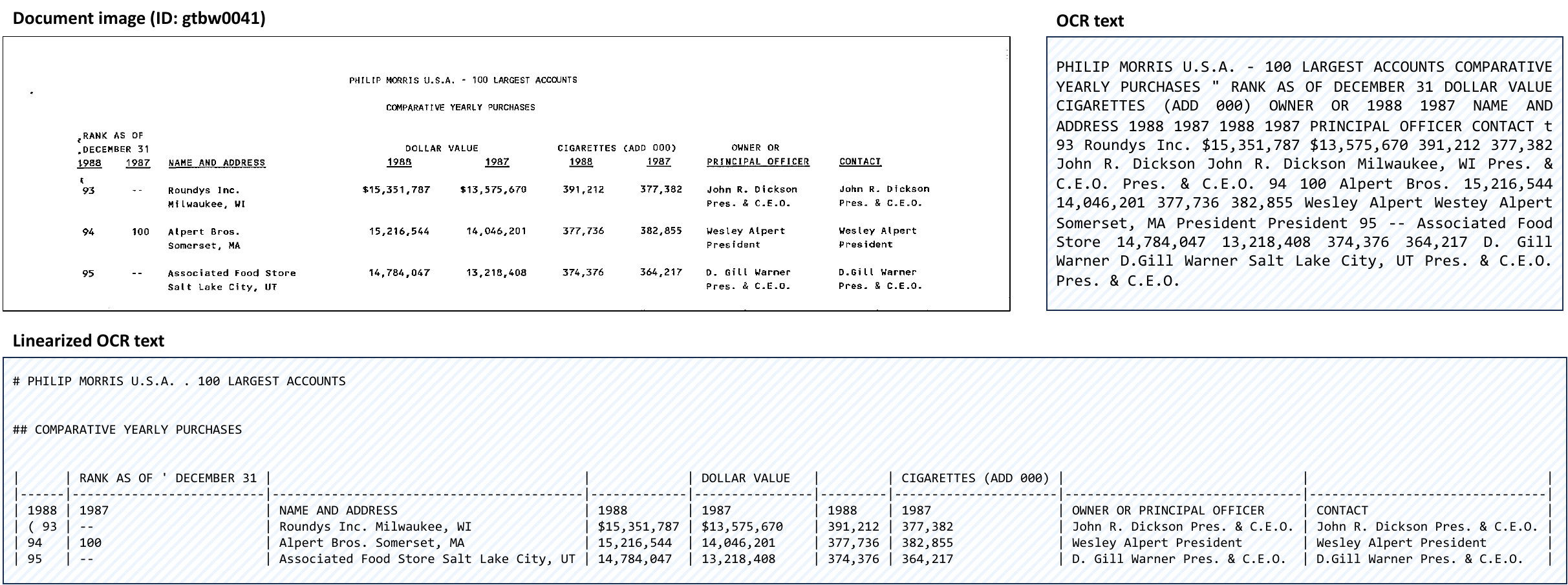}
\end{figure*}
\begin{table*}[!ht]
    \centering
    \scriptsize
    \begin{tabular*}{.8\linewidth}{l}
    \toprule
    KD QAs (using OCR text)\\
    \midrule
    \ttfamily
    Question 1: What is the name of the document? \\
    \ttfamily
    Answer 1: PHILIP MORRIS U.S.A. - 100 LARGEST ACCOUNTS COMPARATIVE YEARLY PURCHASES \\
    \,\\
    \ttfamily
    Question 2: What are the two columns after the name and address? \\
    \ttfamily
    Answer 2: OWNER OR PRINCIPAL OFFICER and CONTACT \\
    \,\\
    \ttfamily
    Question 3: What is the address of Roundys Inc? \\
    \ttfamily
    Answer 3: Milwaukee, WI \\
    \bottomrule
    \toprule
    \textbf{DocKD QAs} (using linearized OCR text) \\
    \midrule
    \ttfamily
    Question 1: What is the title of this document? \\
    \ttfamily
    Answer 1: \# PHILIP MORRIS U.S.A.. 100 LARGEST ACCOUNTS \\
    \,\\
    \ttfamily
    Question 2: What is the name and address of the 94th ranked company in 1988? \\
    \ttfamily
    Answer 2: Alpert Bros. Somerset, MA \\
    \,\\
    \ttfamily
    Question 3: Who is the contact person for Roundys Inc.? \\
    \ttfamily
    Answer 3: John R. Dickson Pres. \& C.E.O. \\
    \bottomrule
    \end{tabular*}
    \vspace{-5pt}
    \caption{Generated QA pairs for the IDL document (ID: gtbw0041).}
    \label{tab:generated_qas-2}
\end{table*}

\paragraph{LLM teachers: Falcon-40B vs. Falcon-180B vs. Claude-2.}
Table\,\ref{tab:generated_qas-3} and Table\,\ref{tab:generated_qas-4} describe the generated QAs from different teacher models, using Falcon-40B, Falcon-180B, and Claude-2. Every teacher utilizes the linaerized OCR text.
The target document in Table\,\ref{tab:generated_qas-3} corresponds to the one used in Table\,\ref{tab:generated_qas-1}, and the document for Table\,\ref{tab:generated_qas-4} corresponds to the one used in Table\,\ref{tab:generated_qas-2}. While Claude-2 adeptly incorporates layout knowledge into QA generation, Falcon-40B tends to produce simple questions and answers, occasionally resulting in duplicates or only slight variations. In contrast, the Falcon-180B model better generates diverse QA pairs, and they are mostly accurate. The primary distinction from Claude-2 lies in the observation that Claude-2 is more inclined to explicitly mention layout information in the document.

\begin{table*}[!h]
    \centering
    \scriptsize
    \begin{tabular*}{.8\linewidth}{l}
    \toprule
    Falcon-40B-generated QAs \\
    \midrule
    \ttfamily
    Question 1: What is the name of the person being questioned? \\
    \ttfamily
    Answer 1: Mr. Howard \\
    \,\\
    \ttfamily
    Question 2: What is the name of the person being questioned? \\
    \ttfamily
    Answer 2: Mr. Howard \\
    \,\\
    \ttfamily
    Question 3: What is the name of the person being questioned? \\
    \ttfamily
    Answer 3: Mr. Howard \\
    \bottomrule
    \toprule
    Falcon-180B-generated QAs  \\
    \midrule
    \ttfamily
    Question 1: What is the name of the person mentioned in the document? \\
    \ttfamily
    Answer 1: Mikulay \\
    \,\\
    \ttfamily
    Question 2: What is the objection mentioned in the document? \\
    \ttfamily
    Answer 2: Objection to the form. \\
    \,\\
    \ttfamily
    Question 3: What is the purpose of outbound telemarketing efforts mentioned in the document? \\
    \ttfamily
    Answer 3: To improve the accuracy of data collection. \\
    \bottomrule
    \toprule
    \textbf{Claude-2-generated QAs}  \\
    \midrule
    \ttfamily
    Question 1: What is the name mentioned in line 1? \\
    \ttfamily
    Answer 1: Mikulay \\
    \,\\
    \ttfamily
    Question 2: What does the questioner ask about in lines 5-8? \\
    \ttfamily
    Answer 2: What the person is doing to decrease the percentage of children who request \\
    \ttfamily
    merchandise by sending in information indicating that they smoke the product. \\
    \,\\
    \ttfamily
    Question 3: What does the respondent say they are constantly improving in lines 11-12? \\
    \ttfamily
    Answer 3: Their problems of data collection. \\
    \bottomrule
    \end{tabular*}
    \vspace{5pt}
    \caption{Generated QAs from different teacher models. The IDL document ftjw0181 in Table\,\ref{tab:generated_qas-1} is provided as the linearized text form.}
    \label{tab:generated_qas-3}
\end{table*}

\begin{table*}[!h]
    \centering
    \scriptsize
    \begin{tabular*}{.8\linewidth}{l}
    \toprule
    Falcon-40B-generated QAs \\
    \midrule
    \ttfamily
    Question 1: What is the name of the company that is ranked 94th in the list? \\
    \ttfamily
    Answer 1: Alpert Bros. \\
    \,\\
    \ttfamily
    Question 2: What is the name of the company that is ranked 95th in the list? \\
    \ttfamily
    Answer 2: Associated Food Store \\
    \bottomrule
    \toprule
    Falcon-180B-generated QAs  \\
    \midrule
    \ttfamily
    Question 1: What is the name of the company ranked 93rd in 1988? \\
    \ttfamily
    Answer 1: Roundys Inc. \\
    \,\\
    \ttfamily
    Question 2: What is the name of the contact person for Roundys Inc. in 1988? \\
    \ttfamily
    Answer 2: John R. Dickson\\
    \bottomrule
    \toprule
    \textbf{Claude-2-generated QAs}  \\
    \midrule
    \ttfamily
    Question 1: What is the title of this document? \\
    \ttfamily
    Answer 1: \# PHILIP MORRIS U.S.A.. 100 LARGEST ACCOUNTS \\
    \,\\
    \ttfamily
    Question 2: What is the name and address of the 94th ranked company in 1988? \\
    \ttfamily
    Answer 2: Alpert Bros. Somerset, MA \\
    \,\\
    \ttfamily
    Question 3: Who is the contact person for Roundys Inc.? \\
    \ttfamily
    Answer 3: John R. Dickson Pres. \& C.E.O. \\
    \bottomrule
    \end{tabular*}
    \vspace{5pt}
    \caption{Generated QAs from different teacher models. The IDL document gtbw0041 in Table\,\ref{tab:generated_qas-2} is provided as the linearized text form. If the last answer surpasses the maximum generation sequence length, the resulting QA pairs consist of only the first two elements.}
    \label{tab:generated_qas-4}
\end{table*}

\paragraph{2-step generation of Q\,$\rightarrow$\,A.}
In QA generation for the document VQA task, we have directed the LLM to simultaneously produce both questions and answers.
This approach aims to ensure consistency with the document contents and establish more accurate relationship between the generated question and its corresponding answer. Alternatively, we explore a 2-step generation process where the LLM initially generates a list of questions and subsequently provides answers for them. 

Table\,\ref{tab:generated_qas-5} and Table\,\ref{tab:generated_qas-6} delineate questions and answers generated by Claude-2, comparing the two distinct generation schemes: 2-step generation and QA simultaneous generation. In Table\,\ref{tab:generated_qas-5}, the target document features a table with limited extractable information. During the first step of question generation, Claude-2 manages to produce questions related to the table headers or the index, yet these remain challenging to answer based on the text. As result, the second step generates random number answers.
Conversely, QA pair simultaneous generation yields better questions and answers, effectively leveraging structural information, \eg, column headers or numbers and ratios listed in the table, and creating easy-to-answer questions from them.

Similar observations are found in Table\,\ref{tab:generated_qas-6}, where the document contains a plot and there is not much information other than the header, axes, and axis labels. In the 2-step generation, questions are formulated regarding the efficiency and percentage of the filtraion, which cannot be addressed using the available document content. The resulting answers include phrases like ``\texttt{not mentioned}'' or ``\texttt{not provided}''. Conversely, QA pair generation produces questions that are easily answerable.

\clearpage

\begin{figure*}[!h]
    \centering
    \includegraphics[width=.9\linewidth]{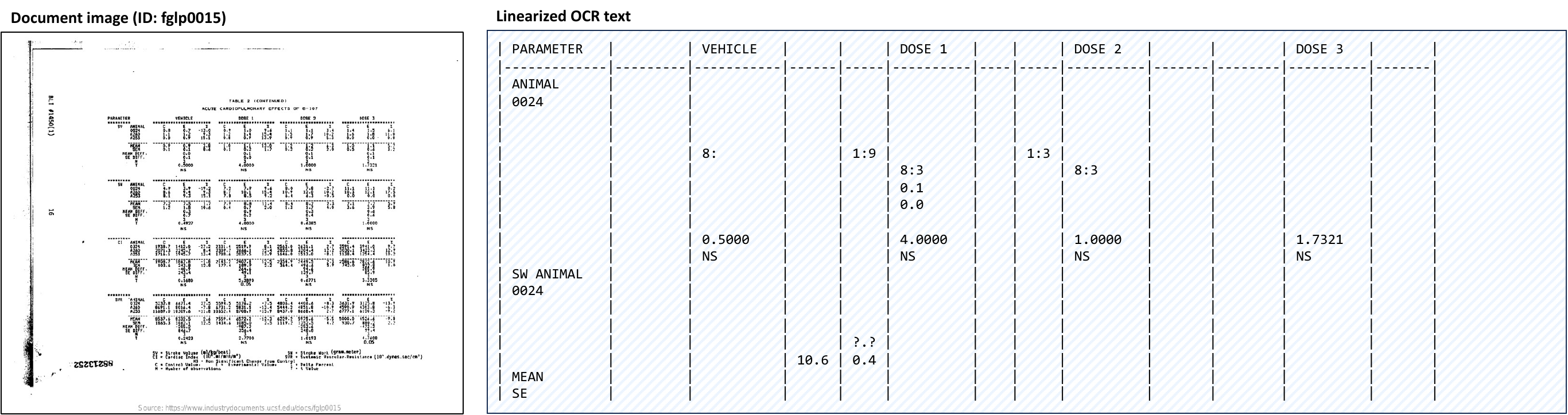}
\end{figure*}
\vspace{-30pt}
\begin{table*}[!h]
    \centering
    \scriptsize
    \begin{tabular*}{.9\linewidth}{l}
    \toprule
    Q\,$\rightarrow$\,A generation \\
    \midrule
    \ttfamily
    Question 1: What is the parameter mentioned in this document? \\
    \ttfamily
    Question 2: What is the animal mentioned in this document? \\
    \ttfamily
    Question 3: What is the dose mentioned in this document? \\
    \midrule
    \ttfamily
    Answer 1: 8: \\
    \ttfamily
    Answer 2: 0024 \\
    \ttfamily
    Answer 3: 8: \\
    \bottomrule
    \toprule
    \textbf{QA pair generation} \\
    \midrule
    \ttfamily
    Question 1: What are the column headers in the table? \\
    \ttfamily
    Answer 1: PARAMETETR, VEHICLE, DOSE 1, DOSE 2, DOSE 3 \\
    \,\\
    \ttfamily
    Question 2: What is the animal number listed in the table? \\
    \ttfamily
    Answer 2: 0024 \\
    \,\\
    \ttfamily
    Question 3: What are the dose ratios listed for Dose 1 and Dose 2? \\
    \ttfamily
    Answer 3: 1:9 and 1:3 \\
    \bottomrule
    \end{tabular*}
    \vspace{5pt}
    \caption{Generated questions and answers for the IDL document (ID: fglp0015).}
    \label{tab:generated_qas-5}
\end{table*}

\begin{figure*}[!h]
    \centering
    \vspace*{5pt}
    \includegraphics[width=.9\linewidth]{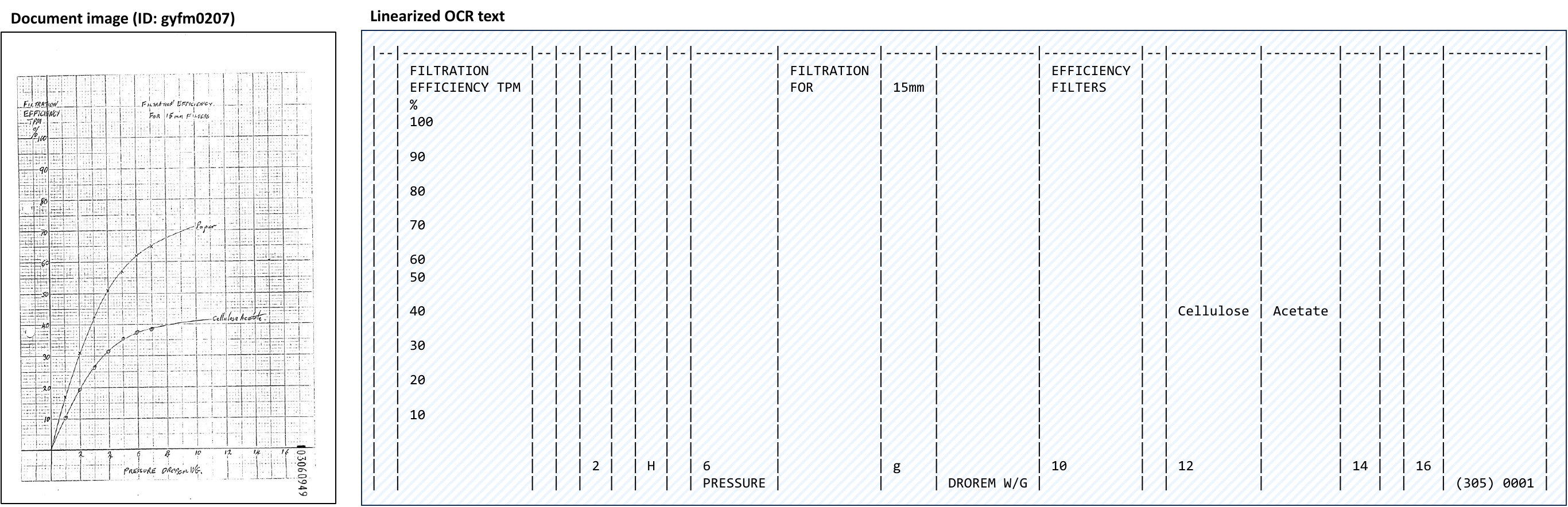}
\end{figure*}
\vspace{-30pt}
\begin{table*}[!h]
    \centering
    \scriptsize
    \begin{tabular*}{.9\linewidth}{l}
    \toprule
    Q\,$\rightarrow$\,A generation \\
    \midrule
    \ttfamily
    Question 1: What is the efficiency of the filtration? \\
    \ttfamily
    Question 2: What is the efficiency of the TPM? \\
    \ttfamily
    Question 3: What is the percentage of the filtration? \\
    \midrule
    \ttfamily
    Answer 1: The efficiency of the filtration is not explicitly mentioned in the given context. \\
    \ttfamily
    Answer 2: The efficiency of the TPM is not mentioned in the given context. \\
    \ttfamily
    Answer 3: The answer to the question is not provided in the given context. \\
    \bottomrule
    \toprule
    \textbf{QA pair generation} \\
    \midrule
    \ttfamily
    Question 1: What is the heading of the document? \\
    \ttfamily
    Answer 1: FILTRATION EFFICIENCY \\
    \,\\
    \ttfamily
    Question 2: What is the abbreviation for the filtration efficiency? \\
    \ttfamily
    Answer 2: TPM \\
    \,\\
    \ttfamily
    Question 3: What is the filter size mentioned in the document? \\
    \ttfamily
    Answer 3: 15mm\\
    \bottomrule
    \end{tabular*}
    \vspace{5pt}
    \caption{Generated questions and answers for the IDL document (ID: gyfm0207).}
    \label{tab:generated_qas-6}
\end{table*}

\clearpage
\subsection{Generated Entities and Fields for Entity Extraction}
\label{subsec:generated_entities}

\begin{figure*}[!ht]
    \minipage{0.48\textwidth}
    \centering
      \includegraphics[height=7.5cm]{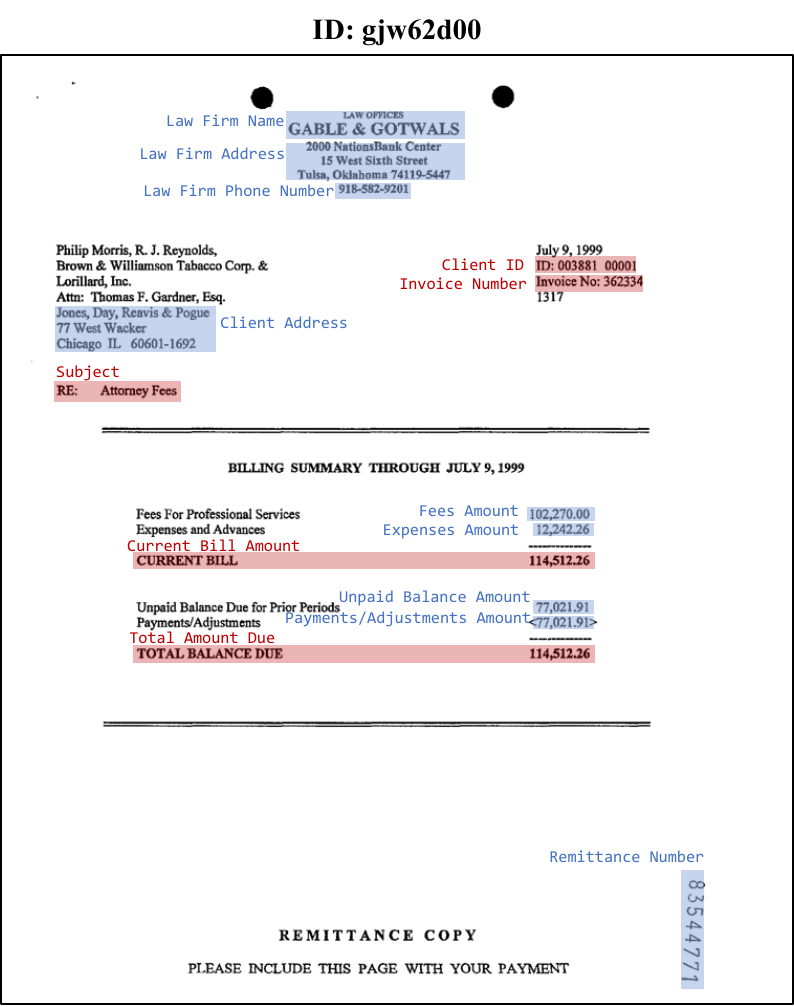}
    \endminipage\hfill
    \minipage{0.48\textwidth}
    \centering
      \includegraphics[height=7.5cm]{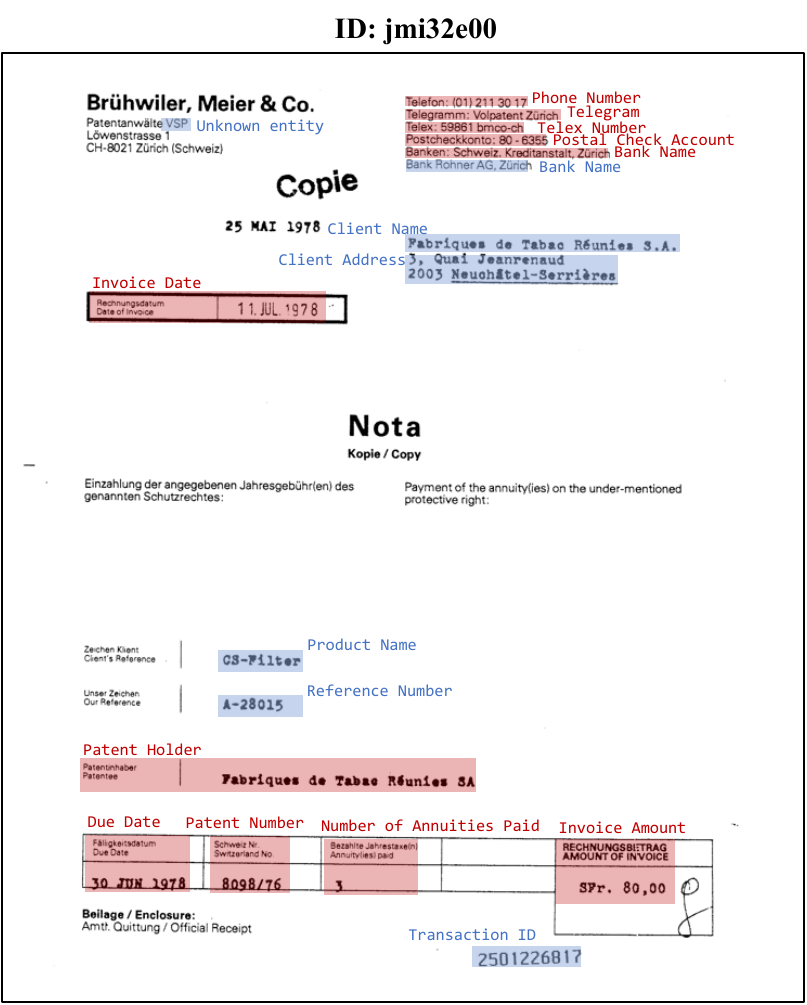}
    \endminipage
  \vskip 15pt
    \minipage{0.51\textwidth}
    \centering
      \includegraphics[height=7.5cm]{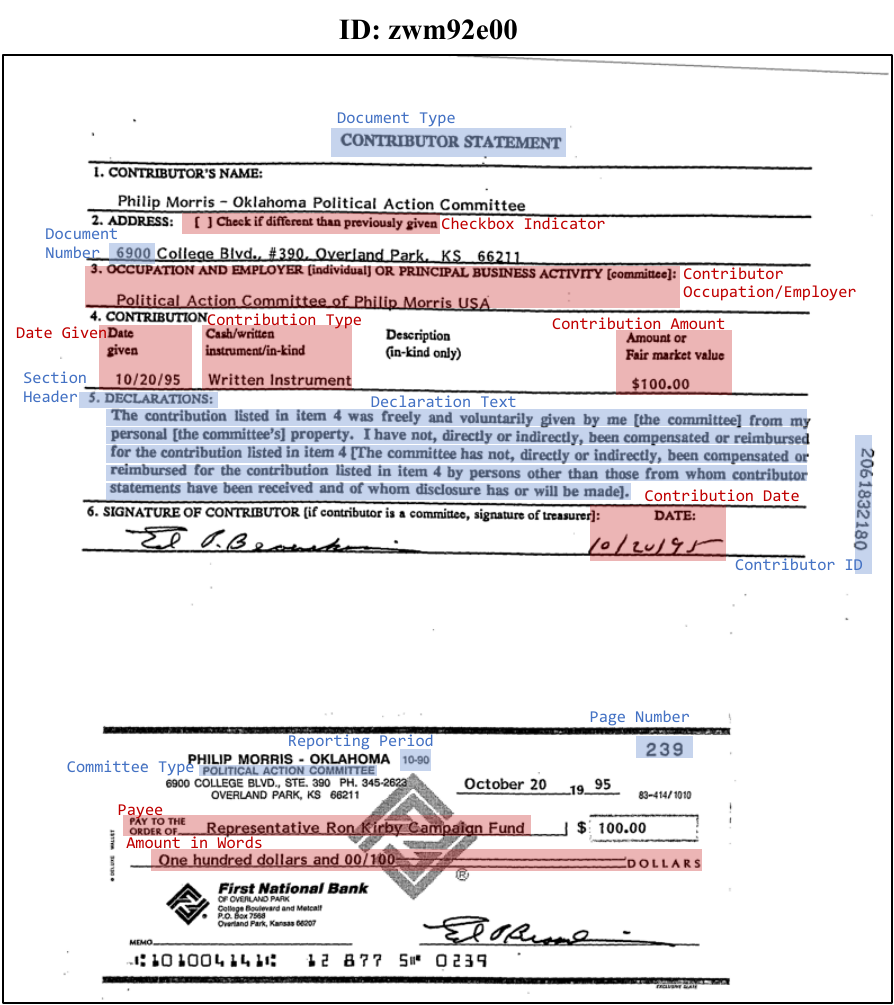}
    \endminipage\hfill
    \minipage{0.47\textwidth}
    \centering
      \includegraphics[height=7.5cm]{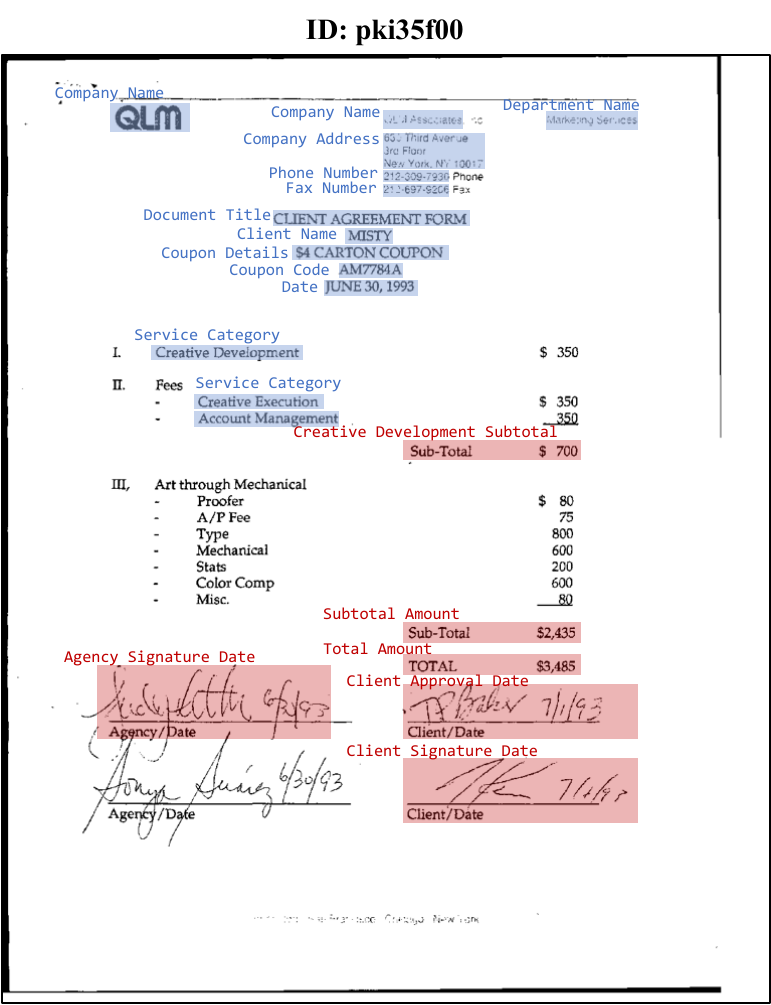}
    \endminipage
  \vskip 15pt
    \minipage{0.47\textwidth}
    \centering
      \includegraphics[height=7.5cm]{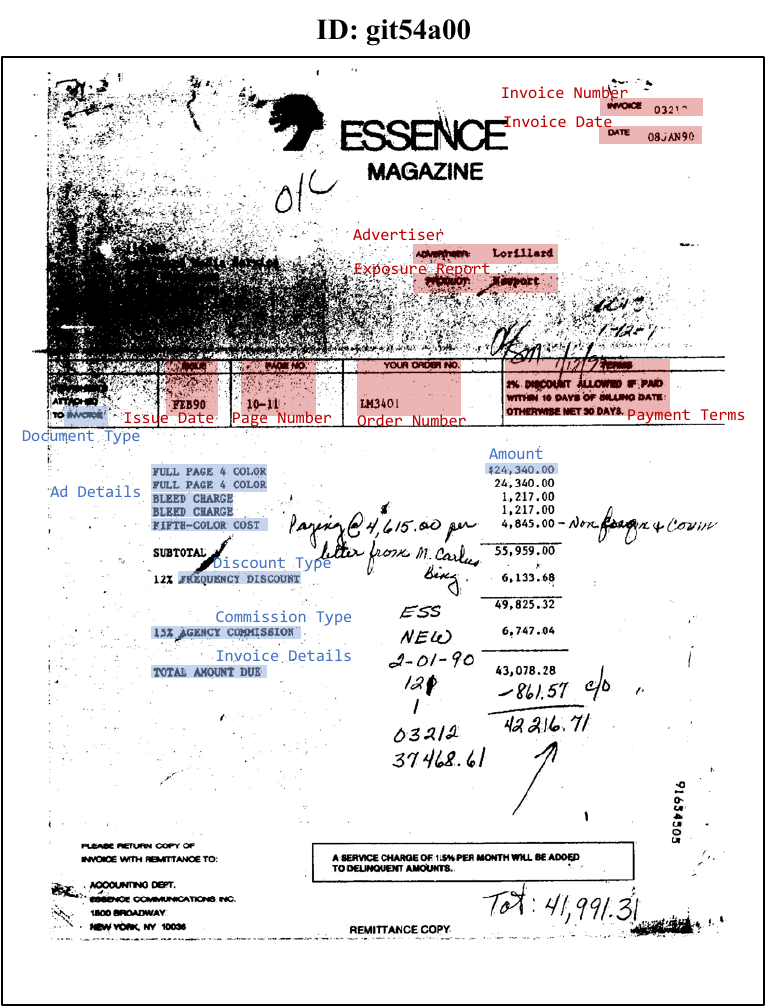}
    \endminipage\hfill
    \minipage{0.51\textwidth}
    \centering
      \includegraphics[height=7.5cm]{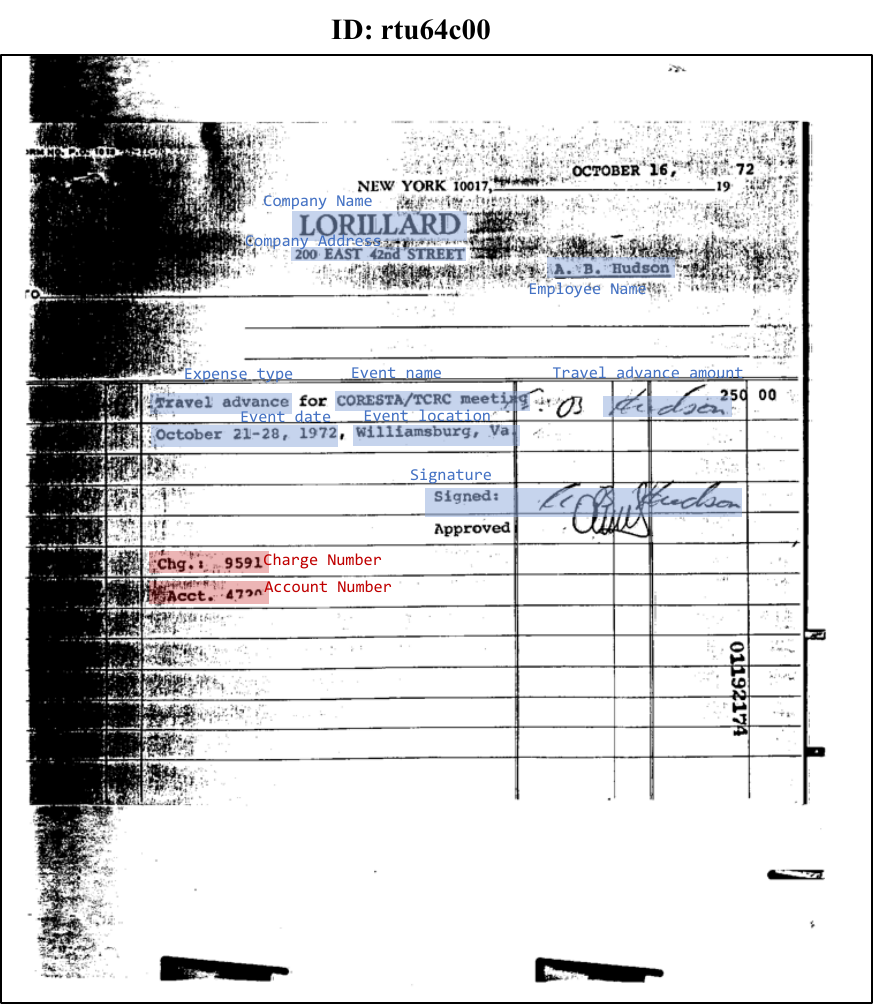}
    \endminipage
    \caption{Generated entities and fields for RVL-CDIP invoice documents.}
    \label{fig:generated_entities-3}
\end{figure*}

\fig\,\ref{fig:generated_entities-3} displays the generated entities and fields for the RVL-CDIP \cite{harley2015rvlcdip} invoice documents. Similar to \fig\,\ref{fig:entity_extraction} in the main paper, non-KV entities and their respective field names are represented by blue boxes and text, while detected KV entities and their corresponding field names are denoted by red boxes and text.
It includes an example where the document is non-English (id: jmi32e00); surprisingly, leveraging the multilingual capability of the LLM, informative entities are extracted and field names are generated in English. Throughout the examples in \fig\,\ref{fig:generated_entities-3}, a diverse range of field names is observed.

Upon generating entities and fields, an aggregation process is employed prior to training the student model. There exist multiple entities within a single document sharing the same field name. We group these entities under the shared field, so that the student model can be trained to match the field to every entity in the group. 
Specifically, we gather all generated field-entity pairs $\{(\mathbf{f}_1, \mathbf{e}_1), (\mathbf{f}_2,\mathbf{e}_2), \dots\}$
and identify the entity group for each field $\mathbf{f}$, $\{\mathbf{e}_j\}$ for all $j$ such that $\mathbf{f}_j=\mathbf{f}$.
Consequently, ${\mathbf{f}}$ is incorporated into $\mathbf{p}_\text{task}$, and ${\{\mathbf{e}_j\}}$ is included in $\mathbf{a}_\text{task}$.

\subsection{Generated Class Labels for Document Classification}
\label{subsec:generated_classes}

\begin{figure*}[!ht]
  \vspace*{10pt}
    \minipage{0.32\textwidth}
    \centering
      \includegraphics[height=11.2cm]{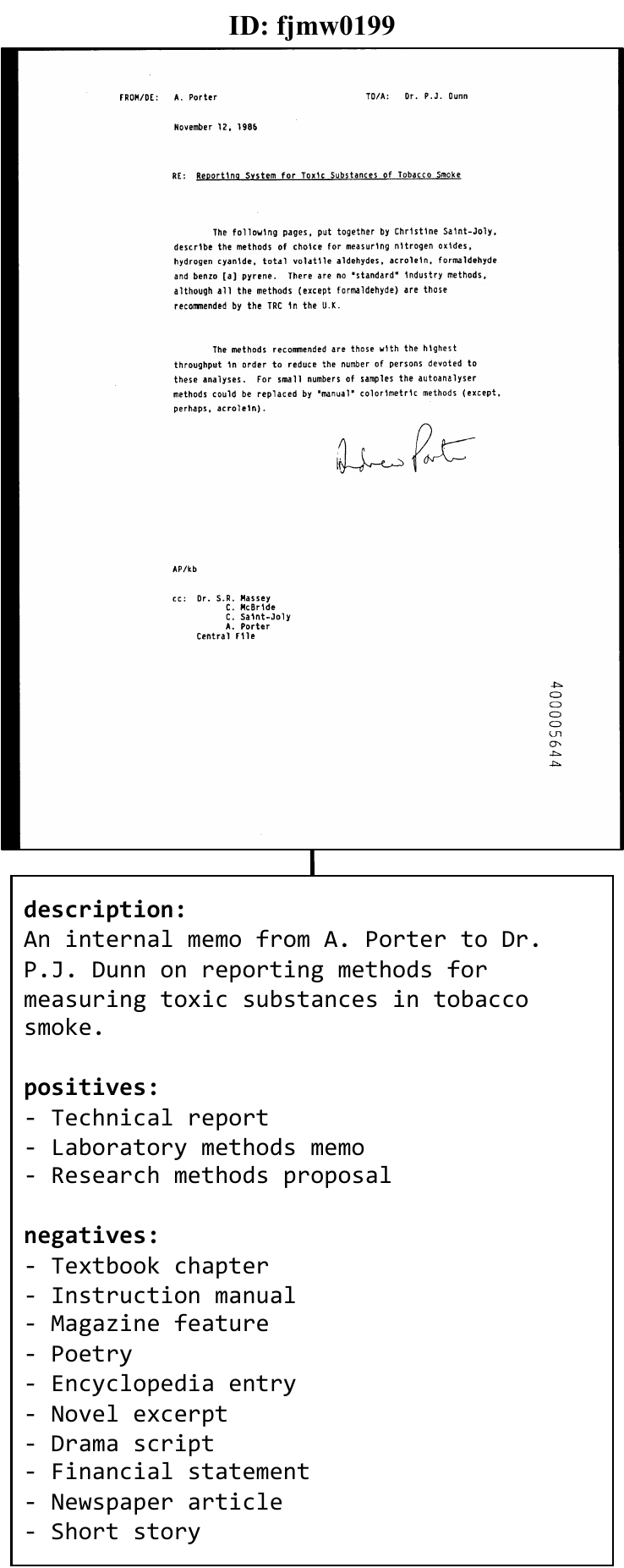}
    \endminipage\hfill
    \minipage{0.32\textwidth}
    \centering
      \includegraphics[height=11.2cm]{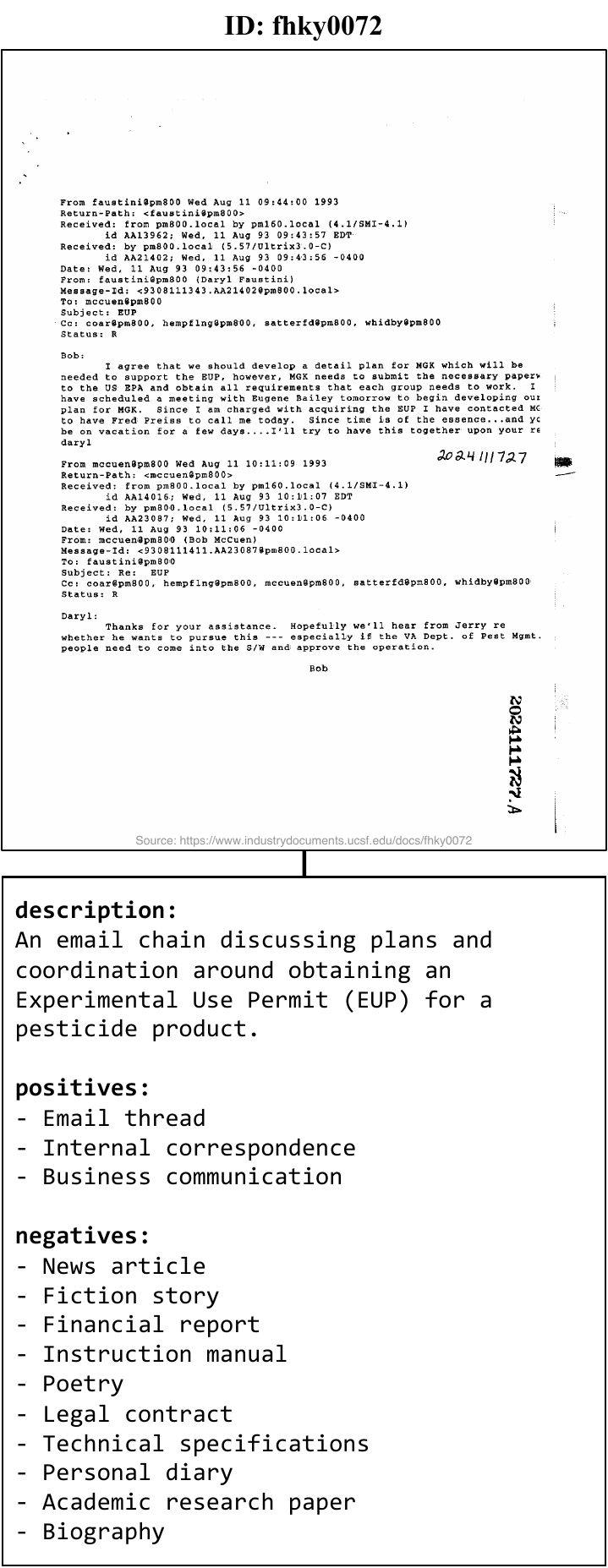}
    \endminipage\hfill
    \minipage{0.32\textwidth}
    \centering
      \includegraphics[height=11.2cm]{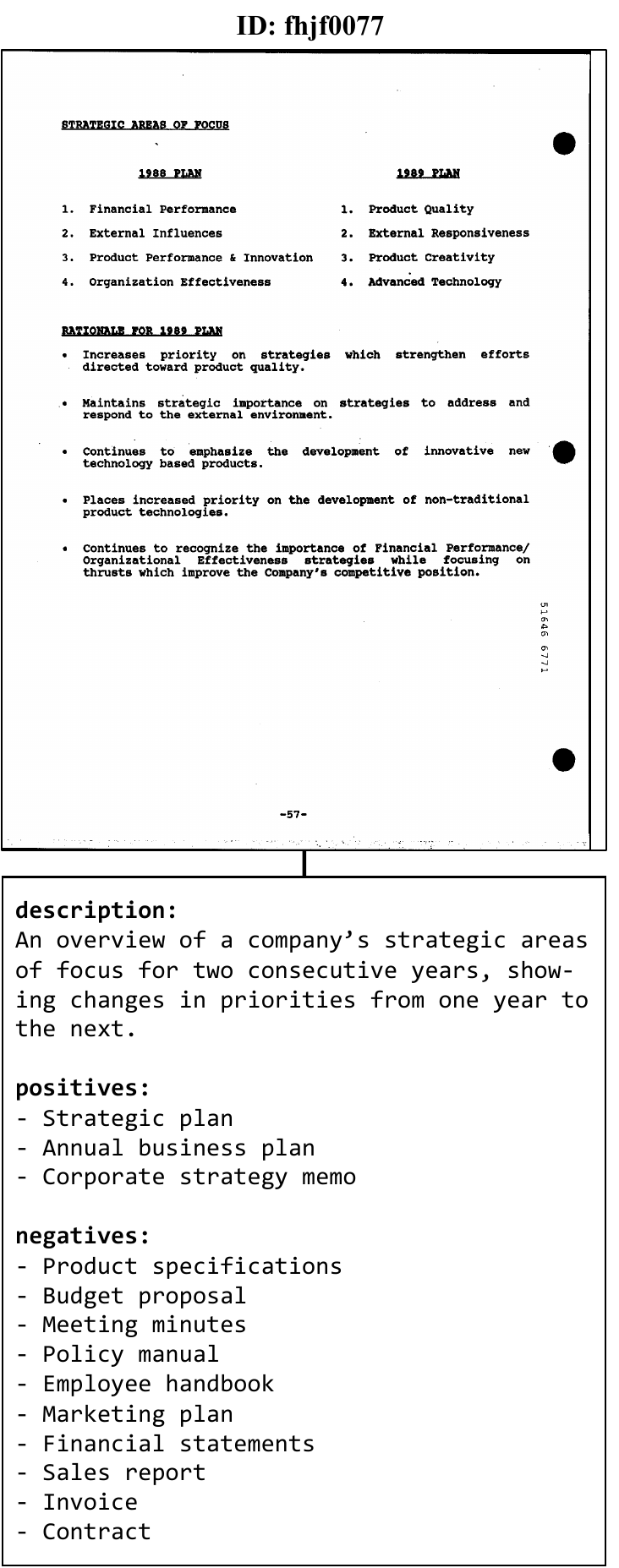}
    \endminipage
  \vskip 20pt
    \minipage{0.32\textwidth}
    \centering
      \includegraphics[height=11.2cm]{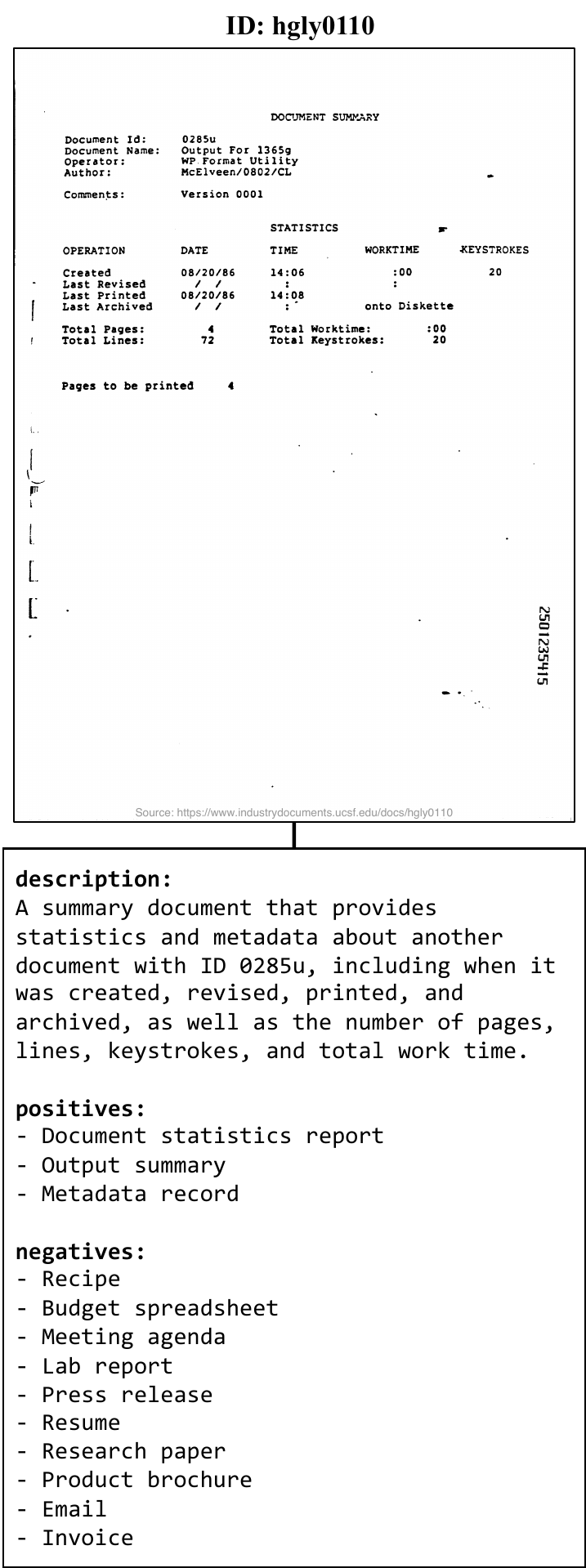}
    \endminipage\hfill
    \minipage{0.32\textwidth}
    \centering
      \includegraphics[height=11.2cm]{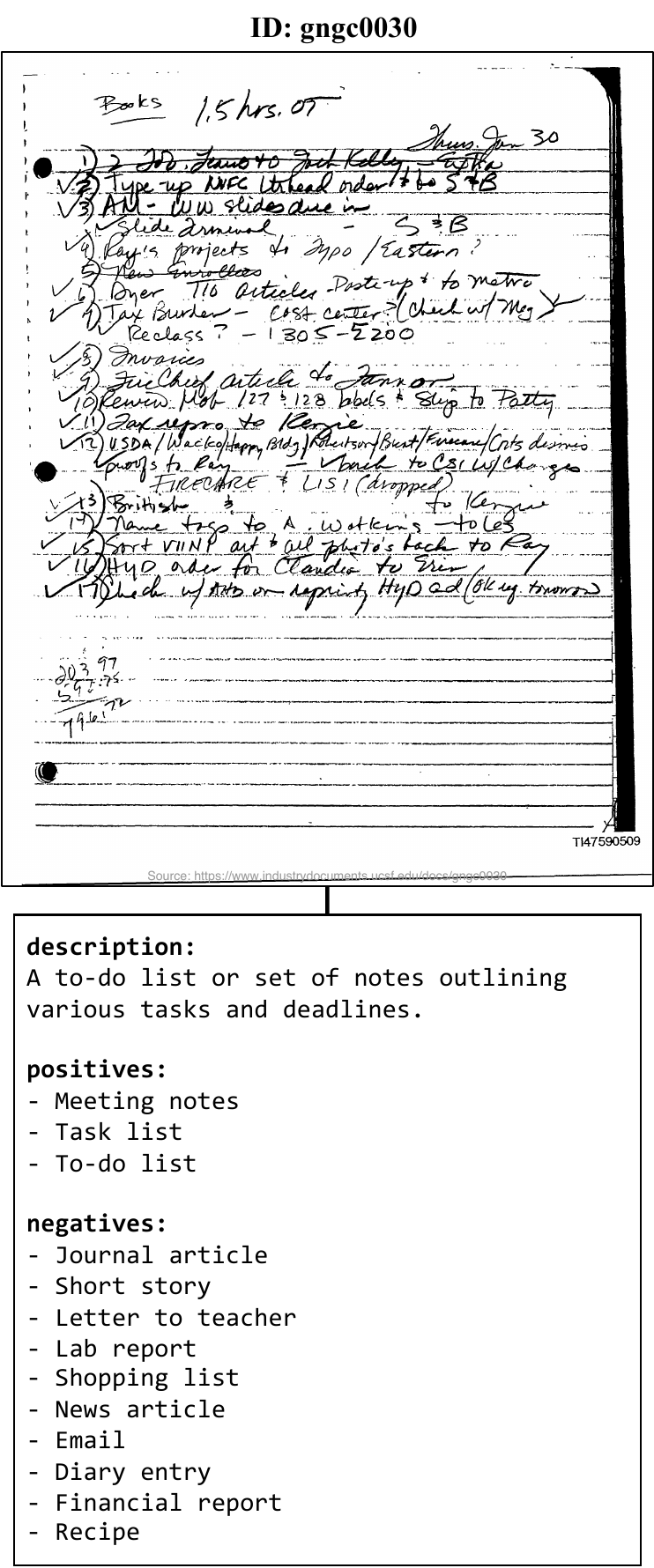}
    \endminipage\hfill
    \minipage{0.32\textwidth}
    \centering
      \includegraphics[height=11.2cm]{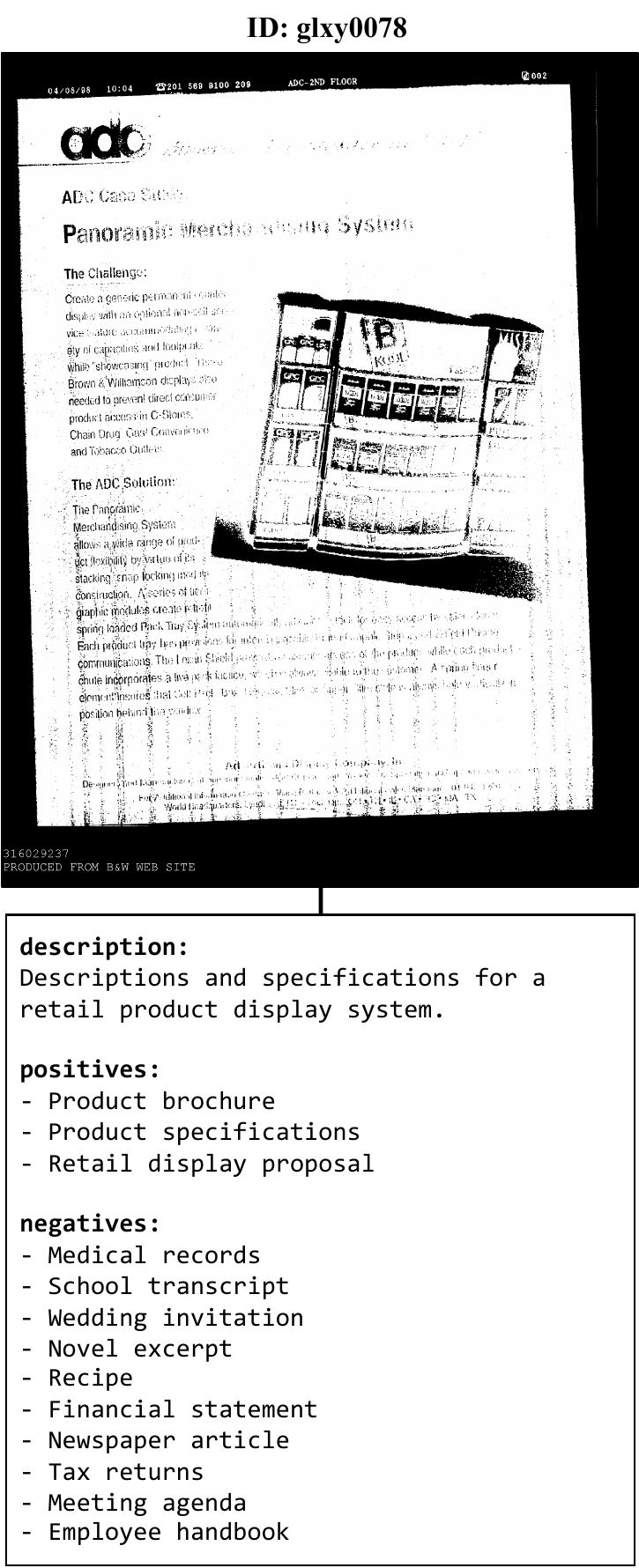}
    \endminipage
    \caption{Generated description and class labels for the IDL documents.}
    \label{fig:generated_classes-2}
\end{figure*}

\fig\,\ref{fig:generated_classes-2} illustrates the generated description, positive class labels, and negative class labels for each IDL \cite{idl} document. The results demonstrate that the LLM generates broad spectrum of class candidates, including report, email, business plan, to-do list, brochure, recipe, poetry, etc. This diversity enables the open document classification capabilities of student models. 

\clearpage
\section{Dataset Specifications}
\label{appx:dataset_specifications}

We provide additional information on the datasets that were not fully described in the main paper.

\paragraph{Evaluation datasets.}
In the document VQA task, we use DocVQA \cite{mathew2021docvqa} as an evaluation dataset. The DocVQA validation set contains manually annotated 5.3K questions related to the real-world industrial documents. For metrics, we use ANLS (average normalized Levenshtein similarity) \cite{biten2019anls} and EM (exact match) which checks if the predicted answer's characters exactly match those of the ground truth.

For the entity extraction, we use two evaluation datasets, CORD \cite{park2019cord} and DeepForm \cite{borchmann2021due}, a collection of restaurant receipts and invoices for political TV ads, respectively. The model should extract entities for the field such as \textit{$<$menu name$>$} or \textit{$<$total cashprice$>$} for CORD, and \textit{$<$advertiser$>$} or \textit{$<$flight to$>$} for DeepForm. The CORD test set is evaluated by entity-level F1 score, while the DeepForm test set is evaluated by ANLS since DeepForm's ground-truth entities are re-formatted from the original document text.

In the classification task, we use RVL-CDIP \cite{harley2015rvlcdip} test set, where 40K documents are labeled into 16 categories, including letter, memo, invoice, form, etc. The performance is measured by the mean accuracy of these 16 categories, while mAcc$^\star$ measures the mean accuracy excluding four ambiguous categories: memo, filefolder, handwritten, and presentation.

\paragraph{Open-set classification.}
In Sec.\,\ref{subsec:analysis}, we have used three out-of-domain datasets for the open-set classification. Here, we outline their setups.
(\textit{i}) RVL-O \cite{larson2022rvlcdip-o} has documents that do not belong to any of 16 categories of RVL-CDIP. These outliers should be classified (or detected) as \textit{other}, with the RVL-CDIP labels also given as candidates.
(\textit{ii}) For IRS-50, we collect 50 types of forms, instructions, and publications from the US Internal Revenue Service.\footnote{\url{https://www.irs.gov/forms-instructions}}
(\textit{iii}) WikiDoc \cite{fujinuma2023multi} consists of 33K Wikipedia screenshots on 111 different subjects.

Table\,\ref{tab:irs50} presents a summary of the 50 IRS class labels which were used in Table\,\ref{tab:document_classification_openset}. Each class label corresponds to one document sample sourced from the US Internal Revenue Service. We also present the precdiction results from Falcon-40B (zero-shot) and DocFormerv2$_\text{base}$ (DocKD).

\begin{table*}[!t]
    \centering
    \small
    \begin{tabular}{l|l|l}
    GT label & Falcon-40B prediction & DFv2$_\text{base}$ S+U prediction \\
    \Xhline{2\arrayrulewidth}
    Form 1000 & Form 1000 & Form 1000 \\
    Form 1040 (Schedule A) & Form 1040 (Schedule A) & \textcolor{red}{Form W-2} \\
    Form 1040 (Schedule B) & Form 1040 (Schedule B) & \textcolor{red}{Form W-2} \\
    Form 1040 (Schedule 1) & Form 1040 (Schedule 1) & \textcolor{red}{Form W-2} \\
    Form 1040 (Schedule 2) & \textcolor{red}{Tax form} & \textcolor{red}{Form W-2} \\
    Form 1040-NR (Schedule NEC) & Form 1040-NR (Schedule NEC) & Form 1040-NR (Schedule NEC) \\
    Form 1040-NR (Schedule OI) & \textcolor{red}{NULL} & \textcolor{red}{Form 1040-NR} \\
    Form 1040-X & \textcolor{red}{Tax form} & Form 1040-X \\
    Form 1098-C & Form 1098-C & Form 1098-C \\
    Form 1098-E & Form 1098-E & Form 1098-E \\
    Form 1098-MA & Form 1098-MA & Form 1098-MA \\
    Form 1098-Q & Form 1098-Q & Form 1098-Q \\
    Form 4506 & Form 4506 & Form 4506 \\
    Form 4506-T & \textcolor{red}{Tax form} & Form 4506-T \\
    Form 4852 & Form 4852 & Form 4852 \\
    Form 8994 & \textcolor{red}{Form} & Form 8994 \\
    Form 9779 & \textcolor{red}{Form} & Form 9779 \\
    Form 9783 & \textcolor{red}{Form 1000} & Form 9783 \\
    Form 15103 & Form 15103 & Form 15103 \\
    Form W-2 & Form W-2 & Form W-2 \\
    Form W-2AS & Form W-2AS & Form W-2AS \\
    Form W-2C & Form W-2C & Form W-2C \\
    Form W-2G & Form W-2G & Form W-2G \\
    Form W-3 & Form W-3 & \textcolor{red}{Form W-2} \\
    Form W-3C & \textcolor{red}{Form W-2C} & \textcolor{red}{Form W-2C} \\
    Form W-3SS & Form W-3SS & \textcolor{red}{Form W-2AS} \\
    Form W-4 & \textcolor{red}{Form 1040 (Schedule 1)} & Form W-4 \\
    Form W-4P & Form W-4P & Form W-4P \\
    Form W-4R & \textcolor{red}{Form 1040 (Schedule 1)} & Form W-4R \\
    Form W-4S & Form W-4S & Form W-4S \\
    Form W-7 & Form W-7 & Form W-7 \\
    Form W-7A & Form W-7A & Form W-7A \\
    Instruction 1040 (Schedule A) & \textcolor{red}{Form 1040 (Schedule A)} & Instruction 1040 (Schedule A) \\
    Instruction 1040 (Schedule B) & \textcolor{red}{Form 1040 (Schedule B)} & \textcolor{red}{Notice 1016} \\
    Instruction 1040-NR & \textcolor{red}{Form} & Instruction 1040-NR \\
    Instruction 1098-Q & Instruction 1098-Q & Instruction 1098-Q \\
    Instruction 8994 & \textcolor{red}{Form 8994} & Instruction 8994 \\
    Notice 1015 & \textcolor{red}{Form 1000} & Notice 1015 \\
    Notice 1016 & \textcolor{red}{Notice} & Notice 1016 \\
    Notice 1027 & \textcolor{red}{Notice} & Notice 1027 \\
    Notice 1392 & \textcolor{red}{Publication} & Notice 1392 \\
    Publication 15 & Publication 15 & Publication 15 \\
    Publication 16 & Publication 16 & Publication 16 \\
    Publication 17 & Publication 17 & Publication 17 \\
    Publication 216 & \textcolor{red}{Publication} & Publication 216 \\
    Publication 1141 & \textcolor{red}{Publication} & Publication 1141 \\
    Publication 1223 & \textcolor{red}{Publication} & Publication 1223 \\
    Publication 1516 & Publication 1516 & Publication 1516 \\
    Publication 1518-A & \textcolor{red}{Publication} & Publication 1518-A \\
    Publication 1546 & \textcolor{red}{Publication} & Publication 1546 \\
    \hline
    \multicolumn{3}{l}{Total count: 50} \\
    \end{tabular}
    \caption{IRS-50 labels and predictions of Falcon-40B and DFv2$_\text{base}$ S+U, which was trained with supervised annotations and unsupervised distillation in Table\,\ref{tab:document_classification_openset}. Red-colored text indicates false predictions.}
    \label{tab:irs50}
\end{table*}

\paragraph{WikiDoc categories.}
The WikiDoc dataset, as described in \citet{fujinuma2023multi}, comprises 111 diverse categories. For each category, the dataset includes screenshots of Wikipedia articles, encompassing a wide range of subjects. Examples of categories in the dataset inlcude Album, BasketballTeam, Cardinal, Dam, Economist, Fish, Glacier, Historian, IceHockeyLeague, Journalist, Lighthouse, Magazine, Noble, OfficeHolder, Poem, Racecourse, School, TradeUnion, University, Volcano, and WrestlingEvent.

\paragraph{DUDE single-page QAs.}
Throughout this paper, our primary focus was on training the student model using single-page document annotations, \ie, document annotation is derived from the contents in a single page. There are document datasets annotated with multi-page information, such as DUDE \cite{borchmann2021due} that is employed for the document VQA task in Table\,\ref{tab:document_vqa_with_dude}. In this case, we only used the QA annotations that can be addressed within a single page.

\end{document}